\documentclass[10pt,twocolumn,letterpaper]{article}
\def\QuantStreamPreprint{1}
\usepackage[pagenumbers]{cvpr}
\usepackage{cuted}
\usepackage{microtype}
\usepackage{multirow}
\usepackage{tabularx}
\usepackage{colortbl}
\usepackage{longtable}
\newcommand{\framework}{QuantStream3D\xspace}
\newcommand{\noqad}{w/o QAD\xspace}
\newcommand{\method}{QuantStream3D\xspace}
\newcommand{\packedbackend}{Packed KV\xspace}
\newcommand{\sg}{\operatorname{sg}}
\graphicspath{{figures/}}
\setlist[itemize]{leftmargin=*,topsep=3pt,itemsep=2pt,parsep=0pt}
\patchcmd{\abstract}{\vspace*{12pt}}{\vspace*{6pt}}{}{\PackageError{QuantStream3D}{Abstract heading spacing patch failed}{Check the current CVPR abstract definition.}}
\patchcmd{\endabstract}{\vspace*{12pt}}{\par\vspace*{4pt}}{}{\PackageError{QuantStream3D}{Abstract ending spacing patch failed}{Check the current CVPR abstract definition.}}
\AtBeginDocument{\hfuzz=0.1pt\vfuzz=0.1pt\hbadness=1000\vbadness=1000}


\newcommand{\analysistakeaway}[2]{%
  \par\smallskip\begingroup\setlength{\fboxsep}{4pt}\setlength{\fboxrule}{0.4pt}%
  \noindent\fcolorbox{cvprblue!55!black}{cvprblue!4}{%
  \begin{minipage}{\dimexpr\linewidth-2\fboxsep-2\fboxrule\relax}%
  \small\textbf{#1} #2\end{minipage}}\par\endgroup\smallskip}

\newcommand{\analysispoints}[2]{%
  \analysistakeaway{#1}{\begin{itemize}[leftmargin=1.1em,topsep=2pt,itemsep=1pt,parsep=0pt]#2\end{itemize}}}

\newcommand{\submissionvspace}[1]{\vspace{#1}}
\newcommand{\paperfigurewidth}[1]{#1\linewidth}

\DeclareMathSizes{13}{12}{9}{6}
\DeclareMathSizes{11}{10.95}{8}{6}
\newcolumntype{Y}[1]{>{\hsize=#1\hsize\linewidth=\hsize\raggedleft\arraybackslash}X}
\definecolor{cvprblue}{rgb}{0.21,0.49,0.74}
\usepackage[pagebackref,breaklinks,colorlinks,allcolors=cvprblue]{hyperref}
\usepackage[capitalize,noabbrev]{cleveref}
\crefname{appendix}{Appendix}{Appendices}
\Crefname{appendix}{Appendix}{Appendices}
\crefname{subappendix}{Appendix}{Appendices}
\Crefname{subappendix}{Appendix}{Appendices}
\crefname{subsubappendix}{Appendix}{Appendices}
\Crefname{subsubappendix}{Appendix}{Appendices}
\NewCommandCopy{\QuantStreamLinkedCref}{\cref}
\RenewDocumentCommand{\cref}{s m}{\QuantStreamLinkedCref{#2}}
\NewCommandCopy{\QuantStreamLinkedCapitalCref}{\Cref}
\RenewDocumentCommand{\Cref}{s m}{\QuantStreamLinkedCapitalCref{#2}}
\def\confName{CVPR}
\def\confYear{2027}
\title{QuantStream3D: Geometry-Aligned Low-Bit KV Cache Quantization\\for Streaming Visual Geometry Transformers}
\makeatletter
\renewcommand{\@maketitle}{%
  \begingroup\centering
    {\Large\bfseries \@title\par}
    \medskip
    {\large \@author\par}
    \medskip
    \begin{minipage}{\textwidth}
\centering
\includegraphics[width=\linewidth]{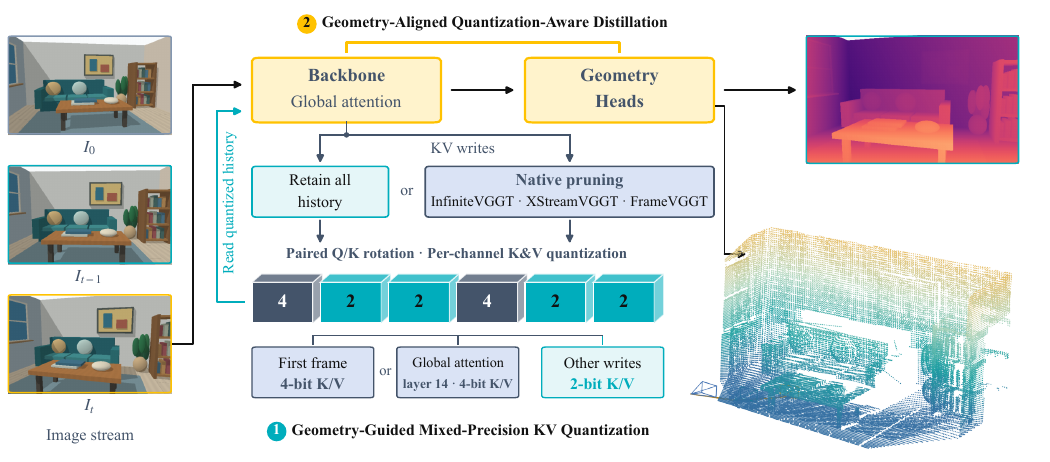}
\captionsetup{hypcap=false}
\captionof{figure}{\textbf{Overview of \framework.} (1) Geometry-Guided Mixed-Precision KV Quantization combines paired Q/K rotation and per-channel K/V quantization. All first-frame writes and all writes at global-attention layer 14 use 4 bits; remaining history uses 2 bits. The policy applies to full history or native-pruning survivors. (2) Geometry-Aligned Quantization-Aware Distillation jointly adapts transformer blocks and geometry heads on student-generated quantized history, supervised by a frozen native-precision teacher.}
\label{fig:teaser}
\captionsetup{hypcap=true}

    \end{minipage}\par\bigskip
  \endgroup}
\AddToHook{cmd/@outputdblcol/before}{%
  \if@firstcolumn\else
    \ifnum\c@page=1
      \setbox\@outputbox\vbox to \ht\@outputbox{%
        \vskip 8pt\unvbox\@outputbox\vss}%
    \fi
  \fi}
\makeatother
\author{\parbox{\dimexpr\textwidth-2\tabcolsep\relax}{\centering\fontsize{11}{14}\selectfont
Zunhai Su\textsuperscript{1}\thanks{Equal contribution.}\quad
Keyu Fan\textsuperscript{2*}\quad
Ruishan Guo\textsuperscript{2}\quad
Zhongzhu Zhou\textsuperscript{3}\quad
Jingzehua Xu\textsuperscript{1}\quad
Bohan Sun\textsuperscript{4}\\
Hansen Feng\textsuperscript{2}\quad
Ruihan Hu\textsuperscript{5}\quad
Yik-Chung Wu\textsuperscript{1}\quad
Ngai Wong\textsuperscript{1}\\
{\normalsize \textsuperscript{1}The University of Hong Kong\quad
\textsuperscript{2}Tsinghua University\quad
\textsuperscript{3}The University of Sydney}\\
{\normalsize \textsuperscript{4}University of Chinese Academy of Sciences\quad
\textsuperscript{5}Harbin Institute of Technology}
}}

\hypersetup{pdftitle={QuantStream3D: Geometry-Aligned Low-Bit KV Cache Quantization for Streaming Visual Geometry Transformers},pdfauthor={Zunhai Su, Keyu Fan, Ruishan Guo, Zhongzhu Zhou, Jingzehua Xu, Bohan Sun, Hansen Feng, Ruihan Hu, Yik-Chung Wu, Ngai Wong},pdfsubject={Preprint}}
\begin{document}
\addtocontents{toc}{\protect\setcounter{tocdepth}{-1}}
\maketitle
\begin{abstract}
Streaming visual geometry transformers retain historical key--value (KV) states as cross-view geometric memory, but the growing cache increasingly limits long-sequence inference. Low-bit quantization reduces its storage cost while altering both the historical evidence and the geometric computation that consumes it. We systematically characterize KV quantization in streaming geometry and uncover two linked challenges: Nonuniform Geometric Sensitivity and Cache--Geometry Mismatch. Geometric sensitivity to KV quantization varies across precision choices and cache locations, while cross-view retrieval perturbations alter geometry-head inputs and propagate through subsequent predictions and KV writes. Guided by these findings, we propose \framework, to our knowledge the first dedicated KV quantization framework for streaming visual geometry transformers. Geometry-Guided Mixed-Precision KV Quantization selectively protects sensitive cross-view evidence, while Geometry-Aligned Quantization-Aware Distillation jointly adapts transformer blocks and geometry heads on student-generated quantized history. Extensive evaluations across five benchmarks and two backbones demonstrate near-lossless geometric accuracy in most settings under predominantly 2-bit KV quantization, with many configurations surpassing their unquantized counterparts. Compared with LLM-oriented KV quantizers, \framework reduces TUM absolute trajectory error and NeuralRGBD completeness error by up to $57.4\%$ and $69.7\%$, respectively. We further integrate \framework with existing native cache-pruning methods, pushing the limits of compression while improving reconstruction quality on NeuralRGBD. Our efficiency analysis shows that our optimized StreamVGGT backend achieves $5.05\times$ physical KV-cache compression and $23.1\%$ lower inference latency at batch size one. Under a fixed memory budget on the same GPU, it supports $5.4\times$ the maximum batch size and delivers up to $1.67\times$ the peak throughput of unquantized inference. \ifdefined\QuantStreamPreprint Code and models: \href{https://github.com/ZunhaiSu/QuantStream3D}{github.com/ZunhaiSu/QuantStream3D}.\else Code is included \mbox{in the supplementary material}.\fi
\end{abstract}

\section{Introduction}
\label{sec:intro}
Feed-forward models now jointly recover camera parameters, depth, and scene structure from images, as exemplified by the Visual Geometry Grounded Transformer (VGGT)~\cite{vggt}. Streaming models such as StreamVGGT and STream3R extend this capability to online inference by reusing historical key--value (KV) states as new views arrive~\cite{streamvggt,stream3r}. This geometric memory carries cross-view evidence into later predictions, while its growing storage and retrieval costs constrain long-sequence inference and concurrency. Recent efforts address this bottleneck through cache pruning, with InfiniteVGGT and FrameVGGT retaining a compact subset of historical states~\cite{infinitevggt,framevggt}. Quantization offers a complementary route by reducing the bit width of retained KV states, an approach widely explored in large language models (LLMs)~\cite{kivi,kvquant,oscar}. Although XStreamVGGT explores combining pruning with quantization, it focuses primarily on 4-bit KV quantization using LLM-oriented methods~\cite{xstreamvggt}. How lower-bit history affects cross-view geometric computation remains insufficiently understood, hindering accurate and efficient KV quantization for streaming geometry.

In this work, we systematically characterize KV quantization across two streaming geometry backbones. Our study spans quantization granularity and bit width, sensitivity across layers, frames, and special tokens, and representation transforms, revealing two linked challenges. \emph{Nonuniform Geometric Sensitivity} spans representation choices and cache locations: K is not uniformly more sensitive than V, challenging LLM-derived key-priority assumptions~\cite{oscar,kvadaquant}. Global attention in layer 14 also forms a shared sensitivity peak across both backbones. \emph{Cache--Geometry Mismatch} arises when quantized cross-view retrieval alters the features supplied to geometry heads trained with native-precision history. Because streaming inference repeatedly consumes and updates this memory, these perturbations can propagate through later queries, KV writes, camera states, and geometric predictions. Together, these findings motivate selective protection of geometric evidence and joint adaptation of history generation, retrieval, and geometric prediction.

We propose \framework (\cref{fig:teaser}) to preserve sensitive geometric memory and align the computation that consumes it. Guided by this analysis, we develop \emph{Geometry-Guided Mixed-Precision KV Quantization}, combining per-channel K/V quantization, selective 2/4-bit precision allocation, and paired Q/K rotation to protect sensitive cross-view evidence. To align geometric computation with this quantized history, we introduce \emph{Geometry-Aligned Quantization-Aware Distillation} (QAD), jointly adapting transformer blocks and geometry heads under a frozen native-precision teacher. The student generates and consumes its own quantized history, allowing later-view geometric supervision to shape earlier KV writes within each clip.

Across five benchmarks and two backbones, many \framework configurations surpass their unquantized counterparts. Relative to LLM-oriented KV quantizers, we reduce absolute trajectory error and reconstruction completeness error by up to $57.4\%$ and $69.7\%$, respectively. We further integrate \framework with existing native cache-pruning methods using the same quantization policy and adapted weights, pushing the limits of compression. To translate low-bit storage into practical inference gains, we implement fused packed-cache kernels for StreamVGGT that reduce intermediate memory traffic and launch overhead.

Our main contributions are summarized as follows:
\begin{itemize}
\item \textbf{Revealing the geometric impact of KV quantization.} We systematically characterize KV quantization in streaming visual geometry transformers, uncovering Nonuniform Geometric Sensitivity and Cache--Geometry Mismatch. We expose the limits of LLM-derived KV quantization assumptions and establish an empirical and analytical basis for geometry-guided quantization and joint adaptation.
\item \textbf{Unifying geometry-guided quantization and geometry-aligned adaptation.} We introduce QuantStream3D, to our knowledge the first dedicated KV quantization framework for streaming visual geometry transformers. It couples Geometry-Guided Mixed-Precision KV Quantization with Geometry-Aligned QAD, preserving sensitive cross-view evidence while jointly adapting transformer blocks and geometry heads to student-generated quantized history.
\item \textbf{Pushing the limits of compression beyond pruning.} We integrate QuantStream3D with three existing native cache-pruning methods, reusing the same quantization policy and adapted weights. This composition achieves up to $6.27\times$ additional KV compression over pruning alone with improved reconstruction quality on NeuralRGBD.
\item \textbf{Delivering geometric accuracy and practical inference efficiency.} We achieve near-lossless or improved geometric accuracy in most settings under predominantly 2-bit KV quantization. At fixed pretrained weights, our fused StreamVGGT backend delivers $5.05\times$ physical KV compression and $23.1\%$ lower inference latency at batch size one. With the same GPU and memory budget, it supports $5.4\times$ the maximum batch size and delivers up to $1.67\times$ the peak throughput of unquantized inference.
\end{itemize}

\section{Related Work}
\label{sec:related}
\subsection{Streaming 3D Geometry}
Feed-forward models infer geometry directly from images: DUSt3R predicts point maps~\cite{dust3r}, while VGGT jointly estimates cameras, depth, and scene structure~\cite{vggt}. StreamVGGT and STream3R extend feed-forward geometry to causal inference, retaining historical KV states for cross-view reasoning~\cite{streamvggt,stream3r}. Recent approaches optimize history through diversity-aware retention (InfiniteVGGT)~\cite{infinitevggt}, coherent memory organization (FrameVGGT)~\cite{framevggt}, spatiotemporal caching (STAC)~\cite{stac}, and pose-guided writes (SceneBound)~\cite{scenebound}.
QVGGT quantizes weights~\cite{qvggt}. Most closely related, XStreamVGGT combines cache pruning with LLM-oriented KIVI quantization, evaluating geometric tasks primarily at 4-bit precision~\cite{xstreamvggt}. Its released quantization path uses simulated quantization and retains dequantized floating-point KV rather than packed low-bit codes. Ultra-low-bit geometric sensitivity and adaptation to quantized history remain insufficiently addressed.
QuantStream3D combines systematic geometric characterization with packed low-bit KV storage and optimized inference kernels.

\subsection{KV Quantization and Adaptation}
KV cache quantization reduces long-context LLM memory costs by storing lower-precision history. KIVI uses per-channel key and per-token value quantization~\cite{kivi}; KVQuant combines per-channel keys, pre-RoPE quantization, and nonuniform formats~\cite{kvquant}. Beyond granularity, per-token methods mitigate outliers through Hadamard rotations (QuaRot)~\cite{quarot} or channel reordering and clipping (SKVQ)~\cite{skvq}. OScaR addresses token-norm imbalance in per-channel quantization using canalized rotation and token scaling~\cite{oscar}, whereas OSCAR derives attention-aware rotations from covariance statistics~\cite{oscarcov}. TurboQuant limits reconstruction and inner-product errors through rotated scalar quantization and residual correction~\cite{turboquant}.
Complementary attention-aware strategies protect salient tokens~\cite{kvsink,akvqvl}. For model adaptation, LLM-QAT uses teacher-generated training data~\cite{llmqat}, while NVFP4 QAD distills a higher-precision teacher~\cite{nvfp4qad}. Streaming geometry requires preserving cross-view evidence and adapting geometric computation to quantized history. QuantStream3D couples geometry-guided KV quantization with Geometry-Aligned QAD, jointly adapting transformer blocks and geometry heads using quantized history generated by the student.

\begin{table*}[t]
\centering\small
\submissionvspace{-5mm}
\caption{K/V granularity and precision in two streaming backbones. Relative changes (\%) use each backbone's unquantized reference: negative indicates improvement; positive indicates degradation. NC denotes normal consistency.}
\label{tab:granularity_precision}
\renewcommand{\arraystretch}{1.06}
\setlength{\tabcolsep}{2pt}
\begin{minipage}[t]{0.48\linewidth}
\centering
\begin{tabular*}{\linewidth}{@{\extracolsep{\fill}}llrrrr@{}}
\toprule
\multicolumn{6}{c}{StreamVGGT \quad Relative change (\%) $\downarrow$}\\
\midrule
\multicolumn{6}{l}{Granularity (K2V2)}\\
\addlinespace[1pt]
Dataset & Metric & KCVC & KCVT & KTVC & KTVT\\
\midrule
Sintel & $\delta_1$ & $+2.3$ & $+2.8$ & $+2.5$ & $+3.5$ \\
KITTI & AbsRel & $+1.0$ & $-6.5$ & $+2.4$ & $-5.2$ \\
Bonn & AbsRel & $+3.9$ & $+3.5$ & $+5.6$ & $+4.4$ \\
TUM & ATE & $+60.2$ & $+83.3$ & $+71.8$ & $+104.3$ \\
NeuralRGBD & NC & $-0.4$ & $+1.3$ & $+0.5$ & $+1.3$ \\
\midrule
\multicolumn{6}{l}{K/V precision (KCVC)}\\
\addlinespace[1pt]
Dataset & Metric & K4V4 & K4V2 & K2V4 & K2V2\\
\midrule
Sintel & $\delta_1$ & $-0.7$ & $+0.5$ & $+2.7$ & $+2.3$ \\
KITTI & AbsRel & $+0.1$ & $+1.2$ & $+2.7$ & $+1.0$ \\
Bonn & AbsRel & $-0.4$ & $+2.7$ & $+0.9$ & $+3.9$ \\
TUM & ATE & $-1.0$ & $+25.4$ & $+15.6$ & $+60.2$ \\
NeuralRGBD & NC & $+0.1$ & $+0.1$ & $-0.8$ & $-0.4$ \\
\bottomrule
\end{tabular*}
\end{minipage}
\hfill
\begin{minipage}[t]{0.48\linewidth}
\centering
\begin{tabular*}{\linewidth}{@{\extracolsep{\fill}}llrrrr@{}}
\toprule
\multicolumn{6}{c}{STream3R \quad Relative change (\%) $\downarrow$}\\
\midrule
\multicolumn{6}{l}{Granularity (K2V2)}\\
\addlinespace[1pt]
Dataset & Metric & KCVC & KCVT & KTVC & KTVT\\
\midrule
Sintel & $\delta_1$ & $+2.8$ & $+3.7$ & $+5.3$ & $+4.2$ \\
KITTI & AbsRel & $+35.7$ & $+43.0$ & $+32.3$ & $+36.5$ \\
Bonn & AbsRel & $+0.1$ & $+0.1$ & $-0.6$ & $+3.4$ \\
TUM & ATE & $+145.3$ & $+135.8$ & $+128.0$ & $+173.2$ \\
NeuralRGBD & NC & $+2.5$ & $+5.4$ & $+5.8$ & $+6.9$ \\
\midrule
\multicolumn{6}{l}{K/V precision (KCVC)}\\
\addlinespace[1pt]
Dataset & Metric & K4V4 & K4V2 & K2V4 & K2V2\\
\midrule
Sintel & $\delta_1$ & $+0.4$ & $+1.5$ & $+1.7$ & $+2.8$ \\
KITTI & AbsRel & $+0.1$ & $+8.3$ & $+10.2$ & $+35.7$ \\
Bonn & AbsRel & $-1.8$ & $-3.0$ & $-4.1$ & $+0.1$ \\
TUM & ATE & $+3.9$ & $+54.8$ & $+47.1$ & $+145.3$ \\
NeuralRGBD & NC & $-0.1$ & $+0.4$ & $+2.7$ & $+2.5$ \\
\bottomrule
\end{tabular*}
\end{minipage}
\end{table*}

\section{\texorpdfstring{\raggedright Characterizing KV Quantization\\*in Streaming 3D Transformers}{Characterizing KV Quantization in Streaming 3D Transformers}}
\label{sec:characterization}
We first introduce streaming inference and KV quantization notation (\cref{sec:preliminaries}). We then examine granularity and K/V precision (\cref{sec:axes}), layer, frame, and token sensitivity (\cref{sec:layers}), and quantization transforms (\cref{sec:transforms}).

\subsection{Preliminaries}
\label{sec:preliminaries}
We use StreamVGGT to illustrate causal geometry inference~\cite{streamvggt,xstreamvggt,infinitevggt}. The image encoder produces patch tokens that, together with camera and register tokens, pass through alternating frame and global attention (GA). Only GA maintains a cross-frame KV cache; frame attention is confined to the current view. For one head at GA layer $\ell$ and frame $t$, let $Q_{\ell,t},K_{\ell,t},V_{\ell,t}$ denote the current-frame tensors, with Q and K defined after normalization and positional encoding. Given head dimension $d$ and historical cache $H_{\ell,t-1}=(K_{\ell,<t},V_{\ell,<t})$, the native-precision GA read is
\begin{equation}
\begin{aligned}
 \bar K_{\ell,t}&=[K_{\ell,<t};K_{\ell,t}],\quad
 \bar V_{\ell,t}=[V_{\ell,<t};V_{\ell,t}],\\
 A_{\ell,t}&=\operatorname{softmax}\!\left(
 Q_{\ell,t}\bar K_{\ell,t}^{\top}/\sqrt d\right)\bar V_{\ell,t}.
\end{aligned}
\label{eq:prelim_attention}
\end{equation}
Here $[\,;\,]$ concatenates tokens. Appending current K/V after each read gives $H_{\ell,t}=(\bar K_{\ell,t},\bar V_{\ell,t})$, so the cache grows with the stream. Geometry heads predict cameras, depth, and point maps; the camera head also maintains recurrent state.

\paragraph{KV quantization and notation.}
We therefore quantize only the historical GA cache $H_{\ell,t-1}$ in \cref{eq:prelim_attention}, decoding it for retrieval while retaining current-frame K/V and camera-head state at native precision. K$a$V$b$ denotes $a$-bit keys and $b$-bit values. Quantization granularity is specified independently of bit width: C and T denote per-channel and per-token quantization. KCVC, KCVT, KTVC, and KTVT denote the channel/channel, channel/token, token/channel, and token/token K/V combinations, respectively. In full-history inference, per-channel groups span tokens at a fixed channel within one frame and head; per-token groups span \mbox{channels within each token}.

\subsection{Quantization Granularity and K/V Precision}
\label{sec:axes}
\label{sec:grouping}

We compare four K/V granularities in both backbones using affine round-to-nearest quantization with groups of 64, without protection or transforms. Both K and V use 2 bits, isolating granularity in this challenging low-bit regime. Per-channel K/V improves trajectory estimation over per-token K/V in both backbones (\cref{tab:granularity_precision}). On StreamVGGT, KCVC also yields lower trajectory error than the LLM-oriented asymmetric KCVT layout adopted by KIVI~\cite{kivi}. Given its strong overall performance, we use KCVC for subsequent analyses unless noted otherwise.

Fixing KCVC, we compare K4V4, K4V2, K2V4, and K2V2 to isolate bit-width and K/V sensitivity. Without protection or transforms, 4-bit caching is comparatively robust, whereas 2-bit caching can substantially increase geometric error (\cref{tab:granularity_precision}). The relative ranking of K4V2 and K2V4 reverses across geometric outputs and backbones. The key-priority tendency reported in LLM studies~\cite{kvadaquant} does not consistently transfer: K is not uniformly more precision-sensitive than V in streaming geometry.

\analysispoints{Takeaways.}{
\item \textbf{Quantization granularity.} Per-channel K/V is a strong overall choice across the evaluated tasks and backbones.
\item \textbf{Bit-width sensitivity.} Unprotected 4-bit caching is comparatively robust; 2-bit caching can incur large geometric errors.
\item \textbf{K/V sensitivity.} K is not consistently more sensitive than V.
}

\subsection{Layer, Frame, and Token Sensitivity}
\label{sec:layers}

\begin{figure}[t]
\centering
\includegraphics[width=\linewidth]{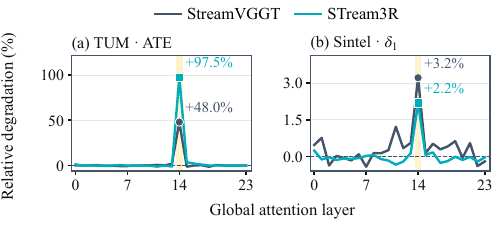}
\caption{\textbf{Layer sensitivity across tasks.} Joint K/V precision drops from 4 to 2 bits at one global-attention layer; all others remain at 4 bits. Relative degradation is measured from each backbone's all-4-bit reference; decreases in $\delta_1$ are positive. Layer 14 is a shared sensitivity peak across both tasks and backbones.}
\label{fig:study}
\end{figure}

We next examine how quantization sensitivity varies across layers, frames, and token types. We scan all 24 global-attention layers in both backbones, reducing K, V, or both from 4 to 2 bits at one layer while keeping all others at 4 bits. Global attention in layer 14 yields the largest error increase under all three interventions and across all three TUM trajectory metrics in both backbones. Joint K/V sensitivity likewise peaks at this layer on Sintel $\delta_1$ (\cref{fig:study}). Independent clip checks confirm the peak (\cref*{app:roles}).

\label{sec:roles}
To examine historical roles and special-token sensitivity~\cite{attentionsinks}, we separately restore all tokens in the first frame, camera tokens across history, or register tokens across history to native precision under otherwise 2-bit caching. First-frame protection lowers translational and rotational relative pose errors in both backbones. Camera- and register-token protection can also improve geometric outputs, but their benefits vary across backbones and metrics (\cref*{app:roles}).

\analysispoints{Takeaways.}{
\item \textbf{Layer sensitivity.} Global attention in layer 14 is a shared sensitivity peak across both backbones.
\item \textbf{Frame sensitivity.} First-frame protection yields repeatable gains in both backbones.
\item \textbf{Token sensitivity.} Camera and register tokens can benefit from protection, with gains that vary across geometric outputs.
}

\subsection{Quantization Transforms}
\label{sec:transforms}

Transformation-based methods reshape activation distributions to ease low-bit quantization. QuaRot and RotateKV redistribute channel outliers through orthogonal mixing~\cite{quarot,rotatekv}. SmoothQuant rescales activations with compensating weight transforms, while QServe applies reciprocal Q/K scaling through SmoothAttention~\cite{smoothquant,qserve}. We examine whether these benefits transfer to streaming geometry.

Under unprotected 2-bit caching, we compare paired Q/K Hadamard rotation, V-only rotation, their combination, and fixed SmoothQuant-style channel scaling. Paired Q/K rotation reduces all three TUM trajectory errors in both backbones. V rotation can improve individual outputs, but adding it to Q/K rotation increases all three STream3R trajectory errors. Fixed channel scaling likewise offers no consistent benefit across tasks and backbones (\cref*{app:transforms}).

\analysispoints{Takeaways.}{
\item \textbf{Q/K rotation.} Paired Hadamard rotation yields repeatable gains under 2-bit quantization.
\item \textbf{Value rotation.} V rotation can help, with gains that depend on the backbone and accompanying transforms.
\item \textbf{Channel scaling.} Fixed channel scaling does not yield consistent gains across the tested settings.
}

\section{QuantStream3D}
\label{sec:method}
\label{sec:alignment}
\label{sec:spatial}
QuantStream3D preserves sensitive cross-view evidence and aligns geometric computation with quantized history. We describe geometry-guided quantization (\cref{sec:policy}), Geometry-Aligned QAD (\cref{sec:qad}), fused execution (\cref{sec:deployment}), and composition with native pruning (\cref{sec:pruned_encoding}).

\subsection{\texorpdfstring{\raggedright Geometry-Guided Mixed-Precision\\*KV Quantization}{Geometry-Guided Mixed-Precision KV Quantization}}
\label{sec:policy}
\paragraph{Per-channel K/V quantization.}
\label{sec:encoding}
Following \cref{sec:axes}, we partition K and V independently into contiguous blocks of $G=64$ tokens within each frame and global-attention head. For a block $X\in\mathbb R^{G\times d}$, each column $\mathbf x_c=X_{:,c}\in\mathbb R^G$ forms one quantization group with its own scale $s_c$ and zero point $z_c$. Nonconstant groups use $b$-bit affine quantization:
\begin{equation}
\begin{aligned}
 s_c&=(M_c-m_c)/(2^b-1),\qquad z_c=\operatorname{round}(-m_c/s_c),\\
 \mathbf q_c&=\operatorname{clip}\bigl(\operatorname{round}(\mathbf x_c/s_c)+z_c,0,2^b-1\bigr),\\
 \widehat{\mathbf x}_c&=s_c(\mathbf q_c-z_c),
\end{aligned}
\label{eq:affine}
\end{equation}
where $m_c=\min\mathbf x_c$ and $M_c=\max\mathbf x_c$. We bit-pack the integer codes and store scales and zero points in FP32. Current-frame K/V remain at native precision during attention and are quantized only when appended to history.

\paragraph{Geometry-guided precision allocation.}
For a write at layer $\ell$ and frame $f$, our allocation protects the first frame or global attention in layer 14:
\begin{equation}
 b_{\ell,f}=\begin{cases}
 4,& f=0\ \lor\ \ell=14,\\
 2,& \text{otherwise}.
 \end{cases}
 \label{eq:map}
\end{equation}
The allocation is shared across backbones and datasets, with no task-specific thresholds or online importance estimation. Protecting sparse camera or register tokens can improve some outputs (\cref{sec:roles}), but assigning them separate bit widths would fragment the precision layout within each write. A single bit width per write preserves regular per-channel groups for efficient packing and fused history reads.

\paragraph{Paired query/key rotation.}
Before quantizing keys, we apply the same Hadamard transform to Q and K after normalization and 2D rotary positional encoding (RoPE), leaving V unrotated. With $R_H=H_{64}/8$ and $H_{64}H_{64}^{\top}=64I$, attention scores are preserved in exact arithmetic:
\begin{equation}
 \widetilde Q=QR_H,\quad \widetilde K=KR_H,\quad
 \widetilde Q\widetilde K^\top=QK^\top.
 \label{eq:rotation}
\end{equation}
Current and historical keys share the rotated query basis, eliminating inverse key rotation during retrieval.

\begin{figure*}[t]
\centering
\submissionvspace{-6mm}
\includegraphics[width=\paperfigurewidth{0.85}]{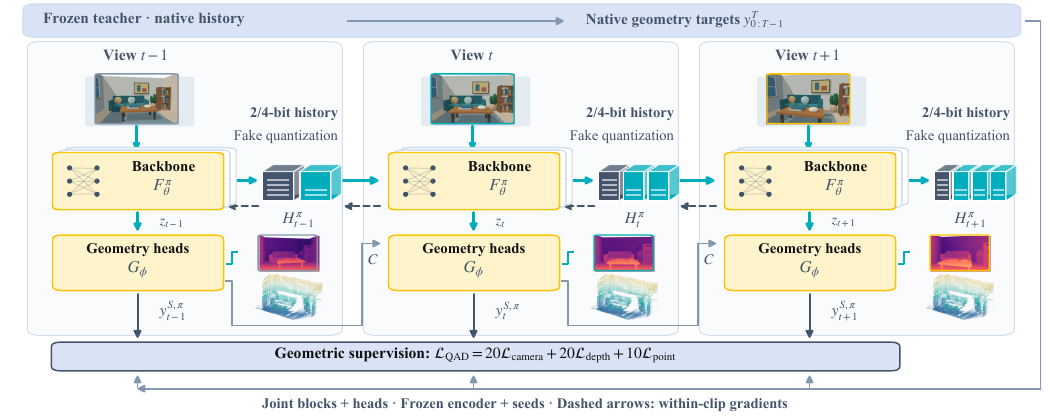}
\caption{\textbf{Geometry-Aligned Quantization-Aware Distillation.} A frozen native-precision teacher supervises joint adaptation of the student's transformer blocks (including FFNs) and geometry heads on student-generated quantized history. Dashed arrows indicate within-clip straight-through gradients through fake-quantized history; $C$ denotes the student's native-precision recurrent camera state.}

\label{fig:qad}
\end{figure*}

\subsection{\texorpdfstring{\raggedright Geometry-Aligned\\*Quantization-Aware Distillation}{Geometry-Aligned Quantization-Aware Distillation}}
\label{sec:qad}
\label{sec:mismatch}
\label{sec:joint}
\paragraph{Cache--geometry mismatch.}
Selective protection preserves sensitive cross-view evidence, but quantized history can still perturb attention retrieval and the features supplied to pretrained geometry heads. At a fixed read $A=PV$, let $P,V$ denote native attention weights and visible values, with current-frame Q/K/V held fixed. Historical-key quantization changes $P$ to $\widehat P$, while historical-value errors $\Delta V$ vanish on current tokens; writing $\widehat A=\widehat P(V+\Delta V)$ and $\Delta P=\widehat P-P$ gives the exact decomposition
\begin{equation}
 \widehat A-A=\Delta P\,V+P\,\Delta V+\Delta P\,\Delta V.
 \label{eq:readout_delta}
\end{equation}
The terms separate attention redistribution, value distortion, and their interaction, linking readout sensitivity to both KV errors and attention context (\cref*{app:readout}). History reuse can carry these perturbations into later queries, KV writes, and geometry-head inputs. With pretrained parameters $(\theta_0,\phi_0)$ fixed, geometry heads predict $y_t=g_{\phi_0}(z_t,C_{t-1})$ from backbone features $z_t$ and unquantized camera state $C_{t-1}$. For native ($0$) and quantized ($\pi$) histories, the geometric output change is approximated to first order by
\begin{equation}
\begin{aligned}
 \Delta y_t
 &=g_{\phi_0}(z_t^\pi,C_{t-1}^\pi)
   -g_{\phi_0}(z_t^0,C_{t-1}^0)\\
 &\simeq J_z\,\Delta z_t+J_C\,\Delta C_{t-1}.
\end{aligned}
\label{eq:head_mismatch}
\end{equation}
Here $\Delta z_t=z_t^\pi-z_t^0$ and $\Delta C_{t-1}=C_{t-1}^\pi-C_{t-1}^0$, with Jacobians $J_z,J_C$ evaluated at the native trajectory. Although $C$ remains unquantized, its values can change through earlier feature perturbations. Paired rollouts on two TUM prefixes show that quantized history changes head inputs, depth, and point predictions at the same view, with pretrained weights and head precision fixed (\cref{tab:head_propagation}).

\begin{table}[t]
\caption{\textbf{Propagation into unquantized geometry heads.} Relative L2 changes (\%) between native- and quantized-history StreamVGGT rollouts at fixed pretrained weights. Head inputs and predictions are paired at the fifth view (\cref*{app:propagation}).}
\label{tab:head_propagation}
\centering\small
\setlength{\tabcolsep}{3.8pt}
\begin{tabularx}{\linewidth}{@{}Xrrr@{}}
\toprule
TUM prefix & Head input & Depth & 3D points\\
\midrule
Sitting halfsphere & 10.24 & 2.17 & 3.13\\
Walking halfsphere & 11.15 & 2.24 & 2.87\\
\bottomrule
\end{tabularx}
\end{table}

\paragraph{Joint geometric supervision.}
To align geometric prediction with quantized history, we propose Geometry-Aligned Quantization-Aware Distillation, jointly adapting history generation, retrieval, and geometric interpretation. A frozen native-precision teacher supervises a student initialized from the same pretrained weights~\cite{llmqat,nvfp4qad}, with independent KV and camera states for each model. The student generates and consumes its own fake-quantized history under the deployment policy. Within-clip straight-through gradients let later-view geometric supervision shape earlier KV writes and subsequent retrieval (\cref{fig:qad}). We jointly update complete frame- and global-attention blocks $\theta$, including FFNs and normalization, and camera, depth, and point heads $\phi$. The image encoder, seed tokens, and bit assignments remain fixed. Let $Y_{\theta,\phi}^{\pi}$ and $Y_{\theta_0,\phi_0}^{0}$ collect student and teacher clip outputs, with stop-gradient $\sg$ on teacher targets; we jointly optimize both parameter groups:
\begin{equation}
 \min_{\theta,\phi}\ \mathbb E_{I_{0:T-1}}
 \mathcal L_{\rm QAD}\!\left(
 Y_{\theta,\phi}^{\pi},\sg(Y_{\theta_0,\phi_0}^{0})\right).
 \label{eq:joint_adaptation}
\end{equation}
Updates to $\theta$ adapt historical writes and retrieval, while updates to $\phi$ adapt geometric interpretation and camera-state recurrence (\cref*{app:joint_scope}). We retain StreamVGGT's released distillation objective and loss weights~\cite{streamvggt}:
\begin{equation}
 \mathcal L_{\rm QAD}=20\mathcal L_{\rm camera}
 +20\mathcal L_{\rm depth}+10\mathcal L_{\rm point}.
 \label{eq:qad}
\end{equation}

\subsection{Fused Execution with Packed KV}
\label{sec:deployment}
Packed KV reduces persistent storage but introduces decoding and input-assembly overhead. We implement fused Triton~\cite{triton} kernels for StreamVGGT to eliminate redundant history materialization and reduce launch overhead~\cite{bitdecoding,oscar}.

On each read, a single kernel unpacks mixed 2/4-bit history, applies affine reconstruction, and copies current native-precision K/V directly into the final contiguous attention operands. Direct placement eliminates per-frame decoded buffers and a separate concatenation step while retaining the original scaled dot-product attention (SDPA) and its autocast behavior. For a combined decoded historical K/V payload of $D_F$ bytes, direct placement avoids $2D_F$ logical bytes of intermediate writes and reads.

We fuse the Q/K Hadamard transforms in a shared launch after normalization and RoPE, leaving V unrotated. Current-frame attention uses native-precision K/V, so each layer stages its writes after the FFN and defers packing until the backbone finishes. Before geometry prediction, one launch per layer jointly performs per-channel min--max reduction, affine quantization, and bit packing for current K/V, while earlier history stays packed (\cref*{app:deployment}).

\subsection{Composition with Native Cache Pruning}
\label{sec:pruned_encoding}
Beyond quantizing unpruned KV caches, we integrate QuantStream3D with existing cache-pruning methods to push the limits of KV compression. Pruning reduces the number of retained states, while quantization lowers their bit width. We instantiate this composition with InfiniteVGGT, XStreamVGGT, and FrameVGGT~\cite{infinitevggt,xstreamvggt,framevggt}, reusing the same adapted StreamVGGT weights. Each integration preserves the native selection rule, selection schedule, auxiliary state, and feature basis. Given decoded KV candidates $Z_\tau$ and selector state $a_\tau$ at event $\tau$, the cache update is
\begin{equation}
\begin{aligned}
 \mathcal I_\tau&=\mathcal S_{\rm base}(Z_\tau,a_\tau),\\
 H_\tau^\pi&=\operatorname{Pack}\!\left[
 \mathcal Q_\pi\!\left(\mathcal G_\pi(Z_\tau[\mathcal I_\tau])\right)\right].
\end{aligned}
\label{eq:pruned_pipeline}
\end{equation}
Here $\mathcal S_{\rm base}$ is the native selector, $\mathcal G_\pi$ forms 64-token groups for each layer, head, and bit width, and $\mathcal Q_\pi$ applies the per-channel quantizer in \cref{sec:policy}. We assign precision using each retained token's original frame and layer identity. Regrouping may combine retained tokens from different frames, so we recompute per-channel min--max statistics for the resulting groups. Quantized history can also change later selection. We therefore evaluate geometric quality and physical KV storage on each realized rollout (\cref{sec:composition}).

\section{Experiments}
\label{sec:experiments}
\begin{table}[t]
\caption{\textbf{Depth estimation on Bonn.} RMSE is in meters; threshold accuracies are percentages. Ours denotes \framework.}
\label{tab:main_bonn}
\centering\small
\renewcommand{\arraystretch}{1.04}
\setlength{\tabcolsep}{2pt}
\begin{tabularx}{\columnwidth}{@{}>{\raggedright\arraybackslash}p{49pt}*{3}{>{\centering\arraybackslash}X}>{\columncolor{black!6}[\tabcolsep][0pt]\centering\arraybackslash}X@{}}
\toprule
Metric & Original & KIVI & QuaRot & Ours\\
\midrule
\multicolumn{5}{c}{\textbf{StreamVGGT}}\\
\midrule
AbsRel$\downarrow$ & 0.059 & 0.061 & 0.063 & 0.055\\
RMSE$\downarrow$ & 0.260 & 0.262 & 0.267 & 0.246\\
$\delta_1$ (\%)$\uparrow$ & 97.2 & 97.4 & 97.4 & 97.5\\
SqRel$\downarrow$ & 0.034 & 0.036 & 0.037 & 0.033\\
LogRMSE$\downarrow$ & 0.114 & 0.115 & 0.116 & 0.108\\
$\delta_2$ (\%)$\uparrow$ & 98.4 & 98.4 & 98.4 & 98.5\\
$\delta_3$ (\%)$\uparrow$ & 99.0 & 99.0 & 99.0 & 99.1\\
\midrule
\multicolumn{5}{c}{\textbf{STream3R}}\\
\midrule
AbsRel$\downarrow$ & 0.070 & 0.070 & 0.070 & 0.062\\
RMSE$\downarrow$ & 0.270 & 0.266 & 0.266 & 0.257\\
$\delta_1$ (\%)$\uparrow$ & 95.2 & 95.3 & 95.0 & 97.4\\
SqRel$\downarrow$ & 0.045 & 0.042 & 0.042 & 0.043\\
LogRMSE$\downarrow$ & 0.133 & 0.130 & 0.130 & 0.119\\
$\delta_2$ (\%)$\uparrow$ & 98.4 & 98.4 & 98.4 & 98.4\\
$\delta_3$ (\%)$\uparrow$ & 98.9 & 98.9 & 99.0 & 98.9\\
\bottomrule
\end{tabularx}
\end{table}

\begin{table}[t]
\caption{\textbf{Camera pose on TUM RGB-D.} Results average eight sequences. ATE and RPE-t are in meters; RPE-r is in degrees.}
\label{tab:main_tum}
\centering\small
\renewcommand{\arraystretch}{1.04}
\setlength{\tabcolsep}{2pt}
\begin{tabularx}{\columnwidth}{@{}>{\raggedright\arraybackslash}p{49pt}*{3}{>{\centering\arraybackslash}X}>{\columncolor{black!6}[\tabcolsep][0pt]\centering\arraybackslash}X@{}}
\toprule
Metric & Original & KIVI & QuaRot & Ours\\
\midrule
\multicolumn{5}{c}{\textbf{StreamVGGT}}\\
\midrule
ATE$\downarrow$ & 0.0269 & 0.0493 & 0.0509 & 0.0264\\
RPE-t$\downarrow$ & 0.0122 & 0.0303 & 0.0297 & 0.0126\\
RPE-r$\downarrow$ & 0.318 & 0.473 & 0.459 & 0.322\\
\midrule
\multicolumn{5}{c}{\textbf{STream3R}}\\
\midrule
ATE$\downarrow$ & 0.0258 & 0.0608 & 0.0509 & 0.0259\\
RPE-t$\downarrow$ & 0.0133 & 0.0369 & 0.0299 & 0.0120\\
RPE-r$\downarrow$ & 0.330 & 0.659 & 0.588 & 0.330\\
\bottomrule
\end{tabularx}
\end{table}

\begin{table}[t]
\caption{\textbf{Reconstruction on NeuralRGBD.} Distances are in meters; NC denotes normal consistency and med. denotes median.}
\label{tab:main_nrgbd}
\centering\small
\renewcommand{\arraystretch}{1.04}
\setlength{\tabcolsep}{2pt}
\begin{tabularx}{\columnwidth}{@{}>{\raggedright\arraybackslash}p{49pt}*{3}{>{\centering\arraybackslash}X}>{\columncolor{black!6}[\tabcolsep][0pt]\centering\arraybackslash}X@{}}
\toprule
Metric & Original & KIVI & QuaRot & Ours\\
\midrule
\multicolumn{5}{c}{\textbf{StreamVGGT}}\\
\midrule
Acc.$\downarrow$ & 0.084 & 0.096 & 0.091 & 0.074\\
Acc. med.$\downarrow$ & 0.044 & 0.048 & 0.049 & 0.043\\
Comp.$\downarrow$ & 0.079 & 0.089 & 0.083 & 0.066\\
Comp. med.$\downarrow$ & 0.039 & 0.040 & 0.043 & 0.033\\
NC$\uparrow$ & 0.861 & 0.850 & 0.843 & 0.867\\
NC med.$\uparrow$ & 0.986 & 0.987 & 0.987 & 0.990\\
NC1$\uparrow$ & 0.862 & 0.861 & 0.845 & 0.876\\
NC2$\uparrow$ & 0.860 & 0.838 & 0.841 & 0.858\\
\midrule
\multicolumn{5}{c}{\textbf{STream3R}}\\
\midrule
Acc.$\downarrow$ & 0.065 & 0.133 & 0.139 & 0.065\\
Acc. med.$\downarrow$ & 0.017 & 0.034 & 0.025 & 0.022\\
Comp.$\downarrow$ & 0.034 & 0.152 & 0.126 & 0.046\\
Comp. med.$\downarrow$ & 0.015 & 0.027 & 0.020 & 0.018\\
NC$\uparrow$ & 0.903 & 0.854 & 0.872 & 0.888\\
NC med.$\uparrow$ & 0.992 & 0.977 & 0.982 & 0.990\\
NC1$\uparrow$ & 0.906 & 0.870 & 0.876 & 0.896\\
NC2$\uparrow$ & 0.900 & 0.838 & 0.869 & 0.881\\
\bottomrule
\end{tabularx}
\end{table}

\begin{table}[t]
\caption{\textbf{Post-pruning reconstruction on NeuralRGBD.} We compare each pruning policy before and after applying QuantStream3D. XStreamVGGT's original KV quantization is disabled in both configurations. Distances are in meters; NC is unitless.}
\label{tab:postprune_neuralrgbd}
\centering\small
\renewcommand{\arraystretch}{1.04}
\setlength{\tabcolsep}{3pt}
\begin{tabularx}{\columnwidth}{>{\raggedright\arraybackslash}p{70pt}*{3}{>{\centering\arraybackslash}X}}
\toprule
Configuration & Acc.$\downarrow$ & Comp.$\downarrow$ & NC$\uparrow$\\
\midrule
\multicolumn{4}{c}{\textbf{Unpruned StreamVGGT}}\\
\midrule
Original & 0.084 & 0.079 & 0.861\\
\midrule
\multicolumn{4}{c}{\textbf{InfiniteVGGT}}\\
\midrule
Prune only & 0.084 & 0.079 & 0.861\\
\addlinespace[1.5pt]
\rowcolor{black!6}
Prune + Ours & 0.074 & 0.066 & 0.867\\
\midrule
\multicolumn{4}{c}{\textbf{XStreamVGGT}}\\
\midrule
Prune only & 0.084 & 0.076 & 0.866\\
\addlinespace[1.5pt]
\rowcolor{black!6}
Prune + Ours & 0.074 & 0.063 & 0.869\\
\midrule
\multicolumn{4}{c}{\textbf{FrameVGGT}}\\
\midrule
Prune only & 0.084 & 0.079 & 0.861\\
\addlinespace[1.5pt]
\rowcolor{black!6}
Prune + Ours & 0.074 & 0.066 & 0.867\\
\bottomrule
\end{tabularx}
\end{table}

\subsection{Experimental Setup}
\label{sec:setup}
We evaluate QuantStream3D on depth estimation (Sintel, KITTI, and Bonn), camera pose estimation (TUM RGB-D), and 3D reconstruction (NeuralRGBD)~\cite{sintel,kitti,bonn,tum,nrgbd}, using StreamVGGT and STream3R. We report three primary metrics per benchmark, totaling 15 per backbone. We compare against the KV-quantization components of KIVI~\cite{kivi} and QuaRot~\cite{quarot} at 2 bits, using affine RTN while keeping weights and other activations at native precision. Baseline configurations and evaluation protocols are detailed in \cref*{app:configuration_ids}. We adapt each backbone for 2,000 updates using ten-view clips from MVS-Synth~\cite{deepmvs} and the PointOdyssey training split~\cite{pointodyssey}. Each adapted checkpoint is shared across all five benchmarks; pruning experiments reuse the adapted StreamVGGT weights. Training details appear in \cref*{app:qad_execution}.

\begin{figure*}[t]
\centering
\submissionvspace{-5mm}
\includegraphics[width=\paperfigurewidth{0.87}]{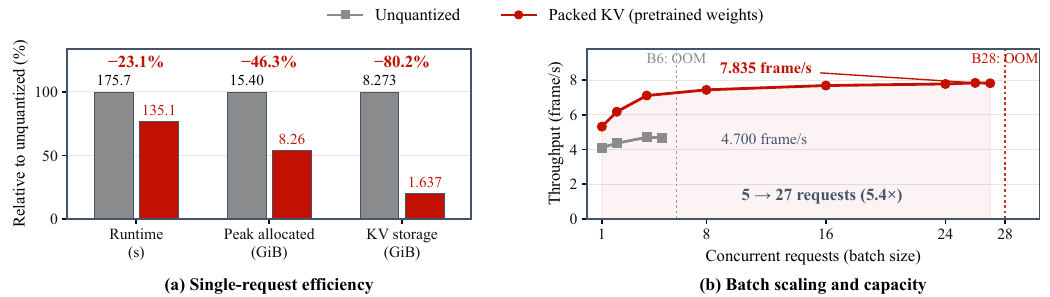}
\caption{\textbf{Complete-inference efficiency on H20.} (a) Batch-size-one runtime (720 views), peak allocated memory, and KV storage relative to original inference (60-GiB budget). (b) Throughput and batch capacity (120-GiB budget). Both paths use pretrained weights and independent 90-view requests. Error bars span three rounds; dashed lines mark adjacent OOM tests.}
\label{fig:efficiency}
\end{figure*}

\subsection{Geometric Accuracy}
\label{sec:recorded}
\label{sec:accuracy}
With predominantly 2-bit history, \framework improves all three primary Bonn metrics over Original, KIVI, and QuaRot on both backbones (\cref{tab:main_bonn}). On TUM RGB-D, it leads both quantized baselines on all three pose metrics across both backbones, reducing STream3R ATE by $57.4\%$ relative to KIVI (\cref{tab:main_tum}). On NeuralRGBD, it improves accuracy, completeness, and normal consistency over both quantized baselines in both backbones; StreamVGGT also improves all three over Original (\cref{tab:main_nrgbd}). For STream3R, completeness error decreases by $69.7\%$ relative to KIVI, although the original model retains lower completeness error and higher normal consistency. Complete benchmark results and QAD controls are provided in \cref*{app:complete_results}.

\subsection{Pushing Compression Beyond Pruning}
\label{sec:composition}
We apply QuantStream3D to InfiniteVGGT, XStreamVGGT, and FrameVGGT, comparing against each pruning-only configuration. Both XStreamVGGT configurations disable its original KV quantization. On NeuralRGBD, QuantStream3D improves all three mean reconstruction metrics for every pruner (\cref{tab:postprune_neuralrgbd}). Completeness error falls by $15.7\%$ for InfiniteVGGT and FrameVGGT, and by $17.5\%$ for XStreamVGGT. QuantStream3D further compresses InfiniteVGGT's retained KV cache by $6.27\times$ relative to pruning alone. \Cref*{app:composition} reports full reconstruction results.

\subsection{Efficiency Analysis}
\label{sec:memory}
\label{sec:fixed}
\begin{table}[t]
\caption{\textbf{Kernel pipeline ablations.} Median GPU time (ms) over five interleaved rounds on fixed synthetic StreamVGGT inputs. The pipeline comprises Q/K rotation, history retrieval, original SDPA, and current K/V encoding. Controls disable one optimization at fixed precision (\cref*{app:kernel_ablation}).}
\label{tab:kernel_ablation}
\centering\small
\renewcommand{\arraystretch}{1.07}
\setlength{\tabcolsep}{4pt}
\begin{tabularx}{\columnwidth}{@{}Xrr@{}}
\toprule
Execution variant & 36-frame history & 89-frame history\\
\midrule
All optimizations & 2.810 & 6.563\\
Without history-read fusion & 7.130 & 17.287\\
Without Q/K pairing & 2.823 & 6.566\\
Without K/V pairing & 2.820 & 6.565\\
\bottomrule
\end{tabularx}
\end{table}

We profile complete StreamVGGT inference on an NVIDIA H20 using fixed pretrained weights, original SDPA, and BF16 autocast (\cref*{app:efficiency}). At batch size one, the packed backend reduces runtime by $23.1\%$, peak allocated memory by $46.3\%$, and persistent KV storage by $80.2\%$ (\cref{fig:efficiency}a). This yields $5.05\times$ physical compression of BF16 KV, including metadata and padding. Under a 120-GiB allocator budget, maximum batch capacity increases from 5 to 27 independent requests ($5.4\times$), bounded by adjacent OOM tests. Peak throughput reaches $1.67\times$ the unquantized baseline (\cref{fig:efficiency}b). With matched synthetic tensors, we isolate history-read fusion, Q/K pairing, and K/V pairing (\cref{tab:kernel_ablation}). Fused reads avoid decoded-frame intermediates and concatenation by reconstructing KV directly in the final attention operands. With 89 historical frames, pipeline latency falls from $17.287$ to $6.563$ ms ($2.63\times$). Pairing reduces isolated Q/K and K/V stage latencies, but paired and separate pipeline timing ranges overlap (\cref*{app:kernel_ablation}).

\subsection{Mechanism Validation and Ablations}
\label{sec:ablations}
\label{sec:mechanism}
\begin{table}[t]
\caption{\textbf{Precision allocation on TUM.} ATE (m) with frozen StreamVGGT and matched final global-attention KV bytes.}
\label{tab:precision_ablation}
\centering\small
\renewcommand{\arraystretch}{1.05}
\setlength{\tabcolsep}{3pt}
\begin{tabularx}{\columnwidth}{@{}l*{4}{>{\centering\arraybackslash}X}@{}}
\toprule
Protected frame & 0 & 1 & 0 & 1\\
Global-attention layer & 14 & 14 & 13 & 13\\
\midrule
ATE$\downarrow$ & 0.0272 & 0.0275 & 0.0347 & 0.0352\\
\bottomrule
\end{tabularx}
\end{table}

\begin{table}[t]
\caption{\textbf{Adaptation scope on NeuralRGBD.} Completeness error (m) with a fixed StreamVGGT cache policy. Adapted variants share initialization, clips, teacher, losses, and 2,000 updates; transformer-block updates include FFNs.}
\label{tab:adaptation_ablation}
\centering\small
\renewcommand{\arraystretch}{1.05}
\setlength{\tabcolsep}{2pt}
\begin{tabularx}{\columnwidth}{@{}l*{3}{>{\centering\arraybackslash}X}>{\columncolor{black!6}[\tabcolsep][0pt]\centering\arraybackslash}X@{}}
\toprule
Scope & \shortstack{No\\adaptation} & \shortstack{Geometry\\heads only} & \shortstack{Transformer\\blocks only} & \shortstack{Joint\\(Ours)}\\
\midrule
Comp.$\downarrow$ & 0.0764 & 0.0760 & 0.0753 & 0.0661\\
\bottomrule
\end{tabularx}
\end{table}
At identical final KV storage, our allocation reduces TUM ATE by $22.6\%$ versus protecting frame 1 and global attention in layer 13 (\cref{tab:precision_ablation}). At a fixed cache policy and 2,000-update budget, jointly adapting transformer blocks and geometry heads reduces NeuralRGBD completeness error by $12.2\%$ over transformer-block adaptation alone (\cref{tab:adaptation_ablation}). Full matched controls and results appear in \cref*{app:ablation_protocol}.

\section{Conclusion}
\label{sec:conclusion}
We systematically characterized low-bit KV quantization in streaming geometry, revealing the need to preserve cross-view evidence and align geometric computation with quantized history. Guided by these findings, QuantStream3D couples geometry-guided mixed precision with Geometry-Aligned QAD. Extensive evaluations demonstrate strong geometric accuracy with predominantly 2-bit history, further compression beyond native pruning, and substantial memory savings and inference gains, providing a practical foundation for efficient streaming 3D perception.

\par\label{page:main_end}
{\raggedright
\bibliographystyle{ieeenat_fullname}
\bibliography{main}

\begin{thebibliography}{36}
\providecommand{\natexlab}[1]{#1}
\providecommand{\url}[1]{\texttt{#1}}
\expandafter\ifx\csname urlstyle\endcsname\relax
  \providecommand{\doi}[1]{doi: #1}\else
  \providecommand{\doi}{doi: \begingroup \urlstyle{rm}\Url}\fi

\bibitem[Ashkboos et~al.(2024)Ashkboos, Mohtashami, Croci, Li, Cameron, Jaggi, Alistarh, Hoefler, and Hensman]{quarot}
Saleh Ashkboos, Amirkeivan Mohtashami, Maximilian~L. Croci, Bo Li, Pashmina Cameron, Martin Jaggi, Dan Alistarh, Torsten Hoefler, and James Hensman.
\newblock {QuaRot}: Outlier-free 4-bit inference in rotated llms.
\newblock \emph{arXiv preprint arXiv:2404.00456}, 2024.

\bibitem[Azinovi{\'c} et~al.(2022)Azinovi{\'c}, Martin-Brualla, Goldman, Nie{\ss}ner, and Thies]{nrgbd}
Dejan Azinovi{\'c}, Ricardo Martin-Brualla, Dan~B Goldman, Matthias Nie{\ss}ner, and Justus Thies.
\newblock Neural {RGB-D} surface reconstruction.
\newblock In \emph{Proceedings of the IEEE/CVF Conference on Computer Vision \mbox{and Pattern Recognition}}, 2022.

\bibitem[Butler et~al.(2012)Butler, Wulff, Stanley, and Black]{sintel}
D.~J. Butler, J. Wulff, G.~B. Stanley, and M.~J. Black.
\newblock A naturalistic open source movie for optical flow evaluation.
\newblock In \emph{European Conference on Computer Vision}, pages 611--625, 2012.

\bibitem[Duanmu et~al.(2024)Duanmu, Yuan, Li, Duan, Zhang, and Lin]{skvq}
Haojie Duanmu, Zhihang Yuan, Xiuhong Li, Jiangfei Duan, Xingcheng Zhang, and Dahua Lin.
\newblock {SKVQ}: Sliding-window key and value cache quantization for large language models.
\newblock \emph{arXiv preprint arXiv:2405.06219}, 2024.

\bibitem[Geiger et~al.(2012)Geiger, Lenz, and Urtasun]{kitti}
Andreas Geiger, Philip Lenz, and Raquel Urtasun.
\newblock Are we ready for autonomous driving? the {KITTI} vision benchmark suite.
\newblock In \emph{Proceedings of the IEEE Conference on Computer Vision \mbox{and Pattern Recognition}}, 2012.

\bibitem[Hariri et~al.(2025)Hariri, Luo, Chen, Zhong, Zhang, Wang, Hu, Han, and Chaudhary]{kvadaquant}
Mohsen Hariri, Alan Luo, Weicong Chen, Shaochen Zhong, Tianyi Zhang, Qifan Wang, Xia Hu, Xiaotian Han, and Vipin Chaudhary.
\newblock Quantize what counts: More for keys, less for values.
\newblock \emph{arXiv preprint arXiv:2502.15075}, 2025.

\bibitem[Hooper et~al.(2024)Hooper, Kim, Mohammadzadeh, Mahoney, Shao, Keutzer, and Gholami]{kvquant}
Coleman Hooper, Sehoon Kim, Hiva Mohammadzadeh, Michael~W. Mahoney, Yakun~Sophia Shao, Kurt Keutzer, and Amir Gholami.
\newblock {KVQuant}: Towards 10 million context length {LLM} inference with {KV} cache quantization.
\newblock In \emph{Advances in Neural Information Processing Systems}, 2024.

\bibitem[Huang et~al.(2018)Huang, Matzen, Kopf, Ahuja, and Huang]{deepmvs}
Po-Han Huang, Kevin Matzen, Johannes Kopf, Narendra Ahuja, and Jia-Bin Huang.
\newblock {DeepMVS}: Learning multi-view stereopsis.
\newblock In \emph{Proceedings of the IEEE Conference on Computer Vision and Pattern Recognition}, pages 2821--2830, 2018.

\bibitem[Lan et~al.(2025)Lan, Luo, Hong, Zhou, Chen, Lyu, Yang, Dai, Loy, and Pan]{stream3r}
Yushi Lan, Yihang Luo, Fangzhou Hong, Shangchen Zhou, Honghua Chen, Zhaoyang Lyu, Shuai Yang, Bo Dai, Chen~Change Loy, and Xingang Pan.
\newblock {STream3R}: Scalable sequential 3d reconstruction with causal transformer.
\newblock \emph{\mbox{arXiv preprint arXiv:2508.10893}}, 2025.

\bibitem[Lin et~al.(2025)Lin, Tang, Yang, Zhang, Xiao, Gan, and Han]{qserve}
Yujun Lin, Haotian Tang, Shang Yang, Zhekai Zhang, Guangxuan Xiao, Chuang Gan, and Song Han.
\newblock {QServe}: {W4A8KV4} quantization and system co-design for efficient {LLM} serving.
\newblock In \emph{Proceedings of Machine Learning and Systems}, 2025.

\bibitem[Liu et~al.(2024{\natexlab{a}})Liu, Oguz, Zhao, Chang, Stock, Mehdad, Shi, Krishnamoorthi, and Chandra]{llmqat}
Zechun Liu, Barlas Oguz, Changsheng Zhao, Ernie Chang, Pierre Stock, Yashar Mehdad, Yangyang Shi, Raghuraman Krishnamoorthi, and Vikas Chandra.
\newblock {LLM-QAT}: Data-free quantization aware training for large language models.
\newblock In \emph{Findings of the Association for Computational Linguistics: ACL}, pages 467--484, 2024{\natexlab{a}}.

\bibitem[Liu et~al.(2024{\natexlab{b}})Liu, Yuan, Jin, Zhong, Xu, Braverman, Chen, and Hu]{kivi}
Zirui Liu, Jiayi Yuan, Hongye Jin, Shaochen Zhong, Zhaozhuo Xu, Vladimir Braverman, Beidi Chen, and Xia Hu.
\newblock {KIVI}: A tuning-free asymmetric 2bit quantization for {KV} cache.
\newblock In \emph{International Conference on Machine Learning}, 2024{\natexlab{b}}.

\bibitem[{OpenBitSys}()]{bitdecoding}
{OpenBitSys}.
\newblock {BitDecoding}.
\newblock \url{https://github.com/OpenBitSys/BitDecoding}.
\newblock Software source revision ae0d836, accessed November 2026.

\bibitem[Palazzolo et~al.(2019)Palazzolo, Behley, Lottes, Giguere, and Stachniss]{bonn}
Emanuele Palazzolo, Jens Behley, Philipp Lottes, Philippe Giguere, and Cyrill Stachniss.
\newblock {ReFusion}: 3d reconstruction in dynamic environments for {RGB-D} cameras exploiting residuals.
\newblock In \emph{IEEE/RSJ International Conference on Intelligent Robots and Systems}, 2019.

\bibitem[Pan et~al.(2026)Pan, Wang, and Wang]{qvggt}
Zhizhen Pan, Hesong Wang, and Huan Wang.
\newblock {QVGGT}: Post-training quantized visual geometry grounded transformer.
\newblock \emph{\mbox{arXiv preprint arXiv:2605.31124}}, 2026.

\bibitem[Sturm et~al.(2012)Sturm, Engelhard, Endres, Burgard, and Cremers]{tum}
Jurgen Sturm, Nikolas Engelhard, Felix Endres, Wolfram Burgard, and Daniel Cremers.
\newblock A benchmark for the evaluation of {RGB-D SLAM} systems.
\newblock In \emph{IEEE/RSJ International Conference on Intelligent Robots and Systems}, 2012.

\bibitem[Su and Yuan(2025)]{kvsink}
Zunhai Su and Kehong Yuan.
\newblock {KVSink}: Understanding and enhancing the preservation of attention sinks in {KV} cache quantization for {LLMs}.
\newblock In \emph{Conference on Language Modeling}, 2025.

\bibitem[Su et~al.(2025{\natexlab{a}})Su, Shen, Li, Chen, Wei, Yu, and Yuan]{akvqvl}
Zunhai Su, Wang Shen, Linge Li, Zhe Chen, Hanyu Wei, Huangqi Yu, and Kehong Yuan.
\newblock {AKVQ-VL}: Attention-aware {KV} cache adaptive 2-bit quantization for vision-language models.
\newblock In \emph{IEEE International Conference on Multimedia and Expo}, pages 1--6, 2025{\natexlab{a}}.

\bibitem[Su et~al.(2025{\natexlab{b}})Su, Wei, Chen, Shen, Li, Yu, and Yuan]{rotatekv}
Zunhai Su, Hanyu Wei, Zhe Chen, Wang Shen, Linge Li, Huangqi Yu, and Kehong Yuan.
\newblock {RotateKV}: Accurate and robust 2-bit {KV} cache quantization for {LLMs} via outlier-aware adaptive rotations.
\newblock In \emph{Proceedings of the Thirty-Fourth International Joint Conference on Artificial Intelligence}, pages 6200--6208, 2025{\natexlab{b}}.

\bibitem[Su et~al.(2026{\natexlab{a}})Su, Yang, Zhang, Liu, Zhang, Wu, Xiong, Du, Zhuang, Qian, Tan, Xie, Wu, Yang, and Wong]{oscar}
Zunhai Su, Rui Yang, Chao Zhang, Yaxiu Liu, Yifan Zhang, Wei Wu, Jing Xiong, Dayou Du, Xialie Zhuang, Yulei Qian, Jianchao Tan, Yuchen Xie, Yik-Chung Wu, Hongxia Yang, and Ngai Wong.
\newblock {OScaR}: The occam's razor for extreme {KV} cache quantization in {LLMs} and beyond.
\newblock \emph{arXiv preprint arXiv:2605.19660}, 2026{\natexlab{a}}.

\bibitem[Su et~al.(2026{\natexlab{b}})Su, Ye, Feng, Fan, Zhang, Yu, Liu, and Wong]{xstreamvggt}
Zunhai Su, Weihao Ye, Hansen Feng, Keyu Fan, Jing Zhang, Dahai Yu, Zhengwu Liu, and Ngai Wong.
\newblock {XStreamVGGT}: Extremely memory-efficient streaming vision geometry grounded transformer with {KV} cache compression.
\newblock \emph{Journal of the Society for Information Display}, 34\penalty0 (5):\penalty0 535--545, 2026{\natexlab{b}}.

\bibitem[Su et~al.(2026{\natexlab{c}})Su, Zhang, Wu, Zhang, Liu, Xiao, Yang, Sun, Yang, Zhang, Xiong, Shen, Fan, Ye, Tao, Wu, Wan, Zhang, Yan, Li, Zhang, Xie, Qian, Xie, Wu, Yang, and Wong]{attentionsinks}
Zunhai Su, Hengyuan Zhang, Wei Wu, Yifan Zhang, Yaxiu Liu, He Xiao, Qingyao Yang, Yuxuan Sun, Rui Yang, Chao Zhang, Jing Xiong, Hui Shen, Keyu Fan, Weihao Ye, Chaofan Tao, Taiqiang Wu, Zhongwei Wan, Tiantian Zhang, Bowen Yan, Zhen Li, Yiming Zhang, Congkai Xie, Yulei Qian, Yuchen Xie, Yik-Chung Wu, Hongxia Yang, and Ngai Wong.
\newblock Attention sink in transformers: A survey on utilization, interpretation, and mitigation.
\newblock \emph{arXiv preprint arXiv:2604.10098}, 2026{\natexlab{c}}.

\bibitem[Tillet et~al.(2019)Tillet, Kung, and Cox]{triton}
Philippe Tillet, H.~T. Kung, and David Cox.
\newblock Triton: An intermediate language and compiler for tiled neural network computations.
\newblock In \emph{Proceedings of the 3rd ACM SIGPLAN International Workshop on Machine Learning and Programming Languages}, pages 10--19, 2019.

\bibitem[Wang et~al.(2025)Wang, Chen, Karaev, Vedaldi, Rupprecht, and Novotny]{vggt}
Jianyuan Wang, Minghao Chen, Nikita Karaev, Andrea Vedaldi, Christian Rupprecht, and David Novotny.
\newblock {VGGT}: Visual geometry grounded transformer.
\newblock In \emph{Proceedings of the IEEE/CVF Conference on Computer Vision \mbox{and Pattern Recognition}}, 2025.

\bibitem[Wang et~al.(2026{\natexlab{a}})Wang, Song, Cai, and Liu]{stac}
Runze Wang, Yuxuan Song, Youcheng Cai, and Ligang Liu.
\newblock {STAC}: Plug-and-play spatio-temporal aware cache compression for streaming 3d reconstruction.
\newblock In \emph{Proceedings of the IEEE/CVF Conference on Computer Vision and Pattern Recognition}, pages 7567--7576, 2026{\natexlab{a}}.

\bibitem[Wang et~al.(2024)Wang, Leroy, Cabon, Chidlovskii, and Revaud]{dust3r}
Shuzhe Wang, Vincent Leroy, Yohann Cabon, Boris Chidlovskii, and Jerome Revaud.
\newblock {DUSt3R}: Geometric 3d vision made easy.
\newblock In \emph{Proceedings of the IEEE/CVF Conference on Computer Vision \mbox{and Pattern Recognition}}, 2024.

\bibitem[Wang et~al.(2026{\natexlab{b}})Wang, Gong, Wang, Zhu, Cao, He, and Li]{scenebound}
Ziming Wang, Chuyue Gong, Chutian Wang, Jingmin Zhu, Hongliang Cao, Yuheng He, and Xiuyu Li.
\newblock {SceneBound}: How much should a streaming 3d reconstruction model remember?
\newblock \emph{\mbox{arXiv preprint arXiv:2611.00756}}, 2026{\natexlab{b}}.

\bibitem[Xiao et~al.(2023)Xiao, Lin, Seznec, Wu, Demouth, and Han]{smoothquant}
Guangxuan Xiao, Ji Lin, Mickael Seznec, Hao Wu, Julien Demouth, and Song Han.
\newblock {SmoothQuant}: Accurate and efficient post-training quantization for large language models.
\newblock In \emph{International Conference on Machine Learning}, 2023.

\bibitem[Xin et~al.(2026)Xin, Priyadarshi, Xin, Kartal, Vavre, et~al.]{nvfp4qad}
Meng Xin, Sweta Priyadarshi, Jingyu Xin, Bilal Kartal, Aditya Vavre, et~al.
\newblock Quantization-aware distillation for {NVFP4} inference accuracy recovery.
\newblock Technical report, NVIDIA, 2026.

\bibitem[Xu and Oishi(2026)]{framevggt}
Zhisong Xu and Takeshi Oishi.
\newblock {FrameVGGT}: Coherence-preserving memory for bounded streaming geometry.
\newblock \emph{\mbox{arXiv preprint arXiv:2603.07690}}, 2026.

\bibitem[Yin et~al.(2019)Yin, Lyu, Zhang, Osher, Qi, and Xin]{steanalysis}
Penghang Yin, Jiancheng Lyu, Shuai Zhang, Stanley Osher, Yingyong Qi, and Jack Xin.
\newblock Understanding straight-through estimator in training activation quantized neural nets.
\newblock In \emph{International Conference on Learning Representations}, 2019.

\bibitem[Yuan et~al.(2026)Yuan, Yang, Yang, Zhang, Zhao, Zhang, and Zhang]{infinitevggt}
Shuai Yuan, Yantai Yang, Xiaotian Yang, Xupeng Zhang, Zhonghao Zhao, Lingming Zhang, and Zhipeng Zhang.
\newblock {InfiniteVGGT}: Visual geometry grounded transformer for endless streams.
\newblock \emph{\mbox{arXiv preprint arXiv:2601.02281}}, 2026.

\bibitem[Zandieh et~al.(2026)Zandieh, Daliri, Hadian, and Mirrokni]{turboquant}
Amir Zandieh, Majid Daliri, Majid Hadian, and Vahab Mirrokni.
\newblock {TurboQuant}: Online vector quantization with near-optimal distortion rate.
\newblock In \emph{International Conference on Learning Representations}, 2026.

\bibitem[Zheng et~al.(2023)Zheng, Harley, Shen, Wetzstein, and Guibas]{pointodyssey}
Yang Zheng, Adam~W. Harley, Bokui Shen, Gordon Wetzstein, and Leonidas~J. Guibas.
\newblock {PointOdyssey}: A large-scale synthetic dataset for long-term point tracking.
\newblock In \emph{Proceedings of the IEEE/CVF International Conference on Computer Vision}, pages 19855--19865, 2023.

\bibitem[Zhou et~al.(2026)Zhou, Zhuang, Li, Chen, Song, Athiwaratkun, and Wu]{oscarcov}
Zhongzhu Zhou, Donglin Zhuang, Jisen Li, Ziyan Chen, Shuaiwen~Leon Song, Ben Athiwaratkun, and Xiaoxia Wu.
\newblock {OSCAR}: Offline spectral covariance-aware rotation for 2-bit {KV} cache quantization.
\newblock \emph{arXiv preprint arXiv:2605.17757}, 2026.

\bibitem[Zhuo et~al.(2025)Zhuo, Zheng, Guo, Wu, Zhou, and Lu]{streamvggt}
Dong Zhuo, Wenzhao Zheng, Jiahe Guo, Yuqi Wu, Jie Zhou, and Jiwen Lu.
\newblock Streaming visual geometry transformer.
\newblock \emph{\mbox{arXiv preprint arXiv:2507.11539}}, 2025.

\end{thebibliography}
}
\clearpage
\onecolumn
\addtocontents{toc}{\protect\setcounter{tocdepth}{2}}
\renewcommand{\contentsname}{Appendix Contents}
\setcounter{tocdepth}{2}
\begingroup
\makeatletter
\renewcommand{\@pnumwidth}{2.8em}
\renewcommand{\@tocrmarg}{3.8em}
\renewcommand{\@dotsep}{8}
\newcommand{\preprinttocentry}[1]{\begingroup\def\suppentry{#1}\patchcmd{\suppentry}{\hskip -1em.~}{}{}{}\suppentry\endgroup}
\renewcommand{\l@section}[2]{\addvspace{7pt}\begingroup\fontsize{13}{19.5}\selectfont\bfseries\@dottedtocline{1}{0em}{1.8em}{\preprinttocentry{#1}}{#2}\endgroup}
\renewcommand{\l@subsection}[2]{\addvspace{1pt}\begingroup\fontsize{12}{18}\selectfont\@dottedtocline{2}{1.8em}{3em}{\preprinttocentry{#1}}{#2}\endgroup}
\makeatother
\tableofcontents
\endgroup
\clearpage
\setcounter{figure}{0}
\setcounter{table}{0}
\setcounter{equation}{0}
\renewcommand{\thefigure}{S\arabic{figure}}
\renewcommand{\thetable}{S\arabic{table}}
\renewcommand{\theequation}{S\arabic{equation}}
\renewcommand{\theHfigure}{supp.\arabic{figure}}
\renewcommand{\theHtable}{supp.\arabic{table}}
\renewcommand{\theHequation}{supp.\arabic{equation}}
\appendix
\crefalias{section}{appendix}
\crefalias{subsection}{appendix}
\crefalias{subsubsection}{appendix}
\section{Additional Quantization Controls}
\label{app:characterization}
\subsection{Layer, Frame, and Token Sensitivity}
\label{app:roles}
We repeat the layer scan on six 32-view clips from MVS-Synth~\cite{deepmvs} and PointOdyssey~\cite{pointodyssey} at $280\times518$, reserving two clips for confirmation. On the confirmation clips, rotation deviations from native predictions again peak at global attention in layer 14 for K-only, V-only, and joint K/V perturbations in both backbones.

We separately restore the first frame, camera tokens, or register tokens to native precision under otherwise 2-bit per-channel K/V caching (\cref{tab:role_protection}). First-frame protection lowers translational and rotational relative pose errors in both backbones. Camera- and register-token protection can also help, but their benefits vary across backbones and outputs. For example, register protection reduces NeuralRGBD completeness error by $13.4\%$ in StreamVGGT but increases it by $3.1\%$ in STream3R.
\begingroup\small
\renewcommand{\arraystretch}{1.04}
\setlength{\tabcolsep}{3pt}
\setlength{\LTpre}{4pt}\setlength{\LTpost}{6pt}\setlength{\LTcapwidth}{\textwidth}
\begin{longtable}{@{\extracolsep{\fill}}lrrrrrrr@{}}
\caption{\textbf{First-frame and special-token protection.} Protected states use native precision; remaining history uses 2-bit per-channel K/V. Controls use 64-value groups without transforms. Distances are in meters and rotations in degrees.}\label{tab:role_protection}\\
\toprule
 & Sintel & KITTI & Bonn & \multicolumn{3}{c}{TUM} & NeuralRGBD\\
\cmidrule(lr){5-7}
Protected states & AbsRel$\downarrow$ & AbsRel$\downarrow$ & AbsRel$\downarrow$ & ATE$\downarrow$ & RPE-t$\downarrow$ & RPE-r$\downarrow$ & Comp.$\downarrow$\\
\midrule
\endfirsthead
\multicolumn{8}{l}{\tablename~\thetable\ (continued)}\\
\toprule
 & Sintel & KITTI & Bonn & \multicolumn{3}{c}{TUM} & NeuralRGBD\\
\cmidrule(lr){5-7}
Protected states & AbsRel$\downarrow$ & AbsRel$\downarrow$ & AbsRel$\downarrow$ & ATE$\downarrow$ & RPE-t$\downarrow$ & RPE-r$\downarrow$ & Comp.$\downarrow$\\
\midrule
\endhead
\bottomrule\endfoot
\multicolumn{8}{l}{\textbf{StreamVGGT}}\\*
None & $0.325$ & $0.174$ & $0.061$ & $0.0431$ & $0.0248$ & $0.428$ & $0.097$\\*
First frame & $0.335$ & $0.179$ & $0.060$ & $0.0439$ & $0.0240$ & $0.408$ & $0.071$\\*
Camera tokens & $0.327$ & $0.174$ & $0.061$ & $0.0435$ & $0.0246$ & $0.430$ & $0.093$\\*
Register tokens & $0.327$ & $0.175$ & $0.061$ & $0.0430$ & $0.0245$ & $0.428$ & $0.084$\\
\midrule
\multicolumn{8}{l}{\textbf{STream3R}}\\*
None & $0.274$ & $0.107$ & $0.070$ & $0.0633$ & $0.0313$ & $0.526$ & $0.141$\\*
First frame & $0.273$ & $0.090$ & $0.067$ & $0.0436$ & $0.0219$ & $0.431$ & $0.139$\\*
Camera tokens & $0.273$ & $0.106$ & $0.069$ & $0.0617$ & $0.0317$ & $0.531$ & $0.141$\\*
Register tokens & $0.274$ & $0.104$ & $0.069$ & $0.0577$ & $0.0281$ & $0.524$ & $0.146$\\
\end{longtable}
\endgroup

\subsection{Quantization Transforms}
\label{app:transforms}
We evaluate Hadamard rotation and fixed channel scaling under unprotected 2-bit per-channel K/V quantization (\cref{tab:transforms}). Paired Q/K rotation reduces all three TUM trajectory errors in both backbones. Adding V rotation reduces NeuralRGBD completeness error by $6.9\%$ in StreamVGGT but increases it by $8.8\%$ in STream3R, where it also worsens all three trajectory errors. Fixed channel scaling yields no consistent improvement across the tested settings.
\begingroup\small
\renewcommand{\arraystretch}{1.04}
\setlength{\tabcolsep}{3pt}
\setlength{\LTpre}{4pt}\setlength{\LTpost}{6pt}\setlength{\LTcapwidth}{\textwidth}
\begin{longtable}{@{\extracolsep{\fill}}lrrrrrrr@{}}
\caption{\textbf{Quantization transforms.} All controls use unprotected 2-bit per-channel K/V with groups of 64. K rotation includes the paired query transform. Distances are in meters and rotation in degrees.}\label{tab:transforms}\\
\toprule
 & Sintel & KITTI & Bonn & \multicolumn{3}{c}{TUM} & NeuralRGBD\\
\cmidrule(lr){5-7}
Transform & AbsRel$\downarrow$ & AbsRel$\downarrow$ & AbsRel$\downarrow$ & ATE$\downarrow$ & RPE-t$\downarrow$ & RPE-r$\downarrow$ & Comp.$\downarrow$\\
\midrule
\endfirsthead
\multicolumn{8}{l}{\tablename~\thetable\ (continued)}\\
\toprule
 & Sintel & KITTI & Bonn & \multicolumn{3}{c}{TUM} & NeuralRGBD\\
\cmidrule(lr){5-7}
Transform & AbsRel$\downarrow$ & AbsRel$\downarrow$ & AbsRel$\downarrow$ & ATE$\downarrow$ & RPE-t$\downarrow$ & RPE-r$\downarrow$ & Comp.$\downarrow$\\
\midrule
\endhead
\bottomrule\endfoot
\multicolumn{8}{l}{\textbf{StreamVGGT}}\\*
No transform & $0.325$ & $0.174$ & $0.061$ & $0.0431$ & $0.0248$ & $0.428$ & $0.097$\\*
K rotation & $0.322$ & $0.175$ & $0.059$ & $0.0358$ & $0.0197$ & $0.378$ & $0.077$\\*
V rotation & $0.331$ & $0.174$ & $0.061$ & $0.0432$ & $0.0253$ & $0.419$ & $0.075$\\*
K + V rotation & $0.325$ & $0.174$ & $0.060$ & $0.0353$ & $0.0203$ & $0.375$ & $0.071$\\*
K scaling & $0.327$ & $0.175$ & $0.061$ & $0.0420$ & $0.0238$ & $0.426$ & $0.093$\\*
V scaling & $0.324$ & $0.174$ & $0.061$ & $0.0430$ & $0.0245$ & $0.431$ & $0.085$\\*
K/V scaling & $0.333$ & $0.176$ & $0.061$ & $0.0439$ & $0.0243$ & $0.428$ & $0.085$\\
\midrule
\multicolumn{8}{l}{\textbf{STream3R}}\\*
No transform & $0.274$ & $0.107$ & $0.070$ & $0.0633$ & $0.0313$ & $0.526$ & $0.141$\\*
K rotation & $0.275$ & $0.088$ & $0.069$ & $0.0426$ & $0.0212$ & $0.445$ & $0.115$\\*
V rotation & $0.274$ & $0.108$ & $0.068$ & $0.0591$ & $0.0304$ & $0.527$ & $0.149$\\*
K + V rotation & $0.268$ & $0.091$ & $0.068$ & $0.0445$ & $0.0214$ & $0.448$ & $0.125$\\*
K scaling & $0.275$ & $0.109$ & $0.069$ & $0.0637$ & $0.0305$ & $0.530$ & $0.141$\\*
V scaling & $0.276$ & $0.110$ & $0.069$ & $0.0646$ & $0.0305$ & $0.533$ & $0.141$\\*
K/V scaling & $0.275$ & $0.107$ & $0.070$ & $0.0649$ & $0.0319$ & $0.530$ & $0.141$\\
\end{longtable}
\endgroup

\section{Geometry-Aligned Quantization-Aware Distillation}
\label{app:qad}
\subsection{Attention Readout Analysis}
\label{app:readout}
To interpret the readout identity in \cref*{eq:readout_delta}, fix a query $q$, the visible tokens, and the attention mask. Let $p=\operatorname{softmax}(Kq/\sqrt d)$ and $a=\sum_i p_i v_i$, with historical errors $e_i^K,e_i^V$ vanishing on current tokens. For $\Delta\ell_i=q^\top e_i^K/\sqrt d$, linearizing the readout gives
\begin{equation}
 \Delta a\simeq\Delta a_{\rm lin}
 =\sum_i p_i\Delta\ell_i(v_i-a)+\sum_i p_i e_i^V.
\label{eq:centered_readout}
\end{equation}
Key errors act through query alignment and value contrast $v_i-a$; value errors are weighted by attention mass $p_i$. Equal tensor-error norms therefore need not produce equal readout changes. With $\Delta p=\widehat p-p$, the exact remainder is
\begin{equation}
 R_a=\Delta a-\Delta a_{\rm lin}
 =V^\top[\Delta p-J(p)\Delta\ell]+\Delta V^\top\Delta p,
 \qquad J(p)=\operatorname{diag}(p)-pp^\top.
\label{eq:readout_remainder}
\end{equation}
The remainder combines nonlinear attention changes and K/V interaction. With operands and support fixed, scaling both perturbations by $\lambda$ gives $R_a=O(\lambda^2)$ as $\lambda\to0$. This forward expansion does not differentiate through rounding.

We evaluate the approximation under the deployment policy on two eight-view TUM Freiburg3 prefixes, \emph{sitting\_halfsphere} and \emph{walking\_halfsphere}. At views 1, 4, and 7 in global-attention layers 0, 14, and 23, we compare native and quantized historical K/V for camera, register, and middle-patch queries, holding current operands fixed. Across these 18 reads, the remainder is $0.93$--$12.10\%$ of the exact change aggregated over the three queries, with a per-query maximum of $24.82\%$. Together, the analysis shows that readout sensitivity depends on both the direction of KV errors and the attention context. We next trace how these readout perturbations propagate into geometric prediction.

\subsection{Propagation into Geometric Predictions}
\label{app:propagation}
\label{app:mechanism_checks}
We compare native- and quantized-history rollouts at fixed parameters, inputs, token identities, and retained observations to trace the propagation in \cref*{eq:head_mismatch}. The states $s_t^0,s_t^\pi$ include decoded history in the shared Q/K basis and native-precision camera state. The residual $\eta_t^\pi$ is the quantized update minus the native update $\mathcal T_t$, both evaluated at $s_{t-1}^\pi$:
\begin{equation}
 s_t^\pi=\mathcal T_t(s_{t-1}^\pi)+\eta_t^\pi,
 \qquad s_t^0=\mathcal T_t(s_{t-1}^0).
\label{eq:state_transition}
\end{equation}
Assume the native transition is $L_t$-Lipschitz on a neighborhood containing both incoming states, with $\|\eta_t^\pi\|\le\epsilon_t$. Starting from the same initial state, the state discrepancy satisfies
\begin{equation}
 \delta_t\le L_t\delta_{t-1}+\epsilon_t
 \le\sum_{i=0}^{t}\epsilon_i\prod_{j=i+1}^{t}L_j,
 \qquad \delta_t=\|s_t^\pi-s_t^0\|.
\label{eq:state_bound}
\end{equation}
The bound follows from the triangle inequality and recursive substitution, with empty products equal to one. Each product bounds the influence of an earlier perturbation through subsequent updates, including camera-state recurrence. It does not prescribe a monotonic trend in geometric prediction error.

We measure propagation on the same TUM prefixes using fixed pretrained weights and independent KV and camera states. Relative L2 changes compare features and predictions at matching views, using the native rollout as the reference. The example in \cref*{tab:head_propagation} pairs feature index 17 with depth and point outputs from the fifth view. Across sampled views, head-input changes at feature indices 4, 11, 17, and 23 range from $0.16$--$11.15\%$, while depth and point outputs change by $1.00$--$2.41\%$ and $1.55$--$3.13\%$, respectively. These paired rollouts confirm that perturbations introduced by quantized history can reach geometric predictions even when geometry heads remain at native precision.

\subsection{Local Capacity of Joint Adaptation}
\label{app:joint_scope}
At fixed parameters, rollout, masks, alignment branch, and confidence weights, we compare the trainable scopes of \cref*{eq:joint_adaptation}. Let $\widetilde r$ be the weighted student--teacher residual and $A_\theta,A_\phi$ the block and head columns of one weighted straight-through output Jacobian, with graph boundaries held fixed. For parameter increments $u,v$, the common linear surrogate is $\widetilde r+A_\theta u+A_\phi v$. Define $A_J=[A_\theta\ A_\phi]$ and $\Pi_X=A_XA_X^\dagger$, where $\dagger$ denotes the pseudoinverse. The unconstrained least-squares optimum is
\begin{equation}
 R_X=\|(I-\Pi_X)\widetilde r\|_2^2,\quad X\in\{\theta,\phi,J\},
 \qquad R_J\le\min\{R_\theta,R_\phi\}.
\label{eq:joint_projection}
\end{equation}
The joint correction space contains both component spaces. Set $r_\theta=(I-\Pi_\theta)\widetilde r$ and $B_\phi=(I-\Pi_\theta)A_\phi$. The columns of $B_\phi$ are head directions orthogonal to the block space, giving
\begin{equation}
 R_\theta-R_J=\|\Pi_{B_\phi}r_\theta\|_2^2,
 \qquad \Pi_{B_\phi}=B_\phi B_\phi^\dagger.
\label{eq:joint_complement}
\end{equation}
The reduction is strict when these directions capture a nonzero component of $r_\theta$: head updates can then complement changes to history generation and retrieval. This is a capacity statement for the shared linear surrogate; the realized gain depends on finite-budget optimization with evolving Jacobians and the original nonquadratic loss. We compare trainable scopes under matched adaptation settings in \cref{tab:scope_complete}.

\subsection{Training Protocol and Gradient Flow}
\label{app:qad_execution}

\begin{table}[!htbp]
\caption{\textbf{Geometry-Aligned QAD settings.} Both backbones use 2,000 updates; cosine horizons follow their released schedules.}
\label{tab:qadrecipe}
\centering\small
\renewcommand{\arraystretch}{1.08}
\begin{tabularx}{\linewidth}{@{}lX@{}}
\toprule
Component & Setting\\
\midrule
Teacher & Frozen native-precision counterpart of each backbone\\
Trainable & Complete frame/global-attention blocks, including FFNs and normalization; camera/depth/point heads\\
Frozen & Image encoder and seed tokens\\
Clip length / global batch & 10 views per clip; 4 clips per update\\
Updates & 2,000\\
Optimizer & AdamW; $\beta=(0.9,0.95)$; weight decay $0.05$\\
Learning rate & $10^{-5}$ peak; $10^{-7}$ floor; 100 warm-up updates\\
Cosine horizon & Original 23,718 updates for StreamVGGT; 16,000 for STream3R\\
Gradients & Norm clipping at 1; no within-clip KV-history detachment\\
Precision & Original BF16 autocast; FP32 loss computation\\
\bottomrule
\end{tabularx}
\end{table}

Adaptation uses ten-view clips from MVS-Synth~\cite{deepmvs} and the PointOdyssey training split~\cite{pointodyssey}, interleaved at 1:5 from pools of 120 and 33 training scenes, respectively. With global batch four, the 2,000-update schedule processes 8,000 training clips.

Teacher and student start from the same pretrained weights and maintain separate KV and camera states. The student generates and consumes its own fake-quantized history under the inference policy in \cref*{sec:policy}. The KV history remains connected to the computation graph throughout the clip; histories and graphs reset between clips, while the original geometry-head graph boundaries are preserved. For fake quantization, we use the affine quantizer with an identity straight-through estimator (STE)~\cite{steanalysis}, holding recomputed min--max statistics and bit assignments fixed during backpropagation.

Let $s_t=\mathcal T_t(s_{t-1};\psi,I_t,\pi)$ collect KV and camera state, with trainable $\psi=(\theta,\phi)$. Under the STE backward map, its parameter sensitivity $J_t=D_\psi^{\rm STE}s_t$ satisfies
\begin{equation}
 J_t=D_{\psi}^{\rm STE}\mathcal T_t+D_s^{\rm STE}\mathcal T_t\,J_{t-1},
 \qquad J_{-1}=0.
\label{eq:history_gradient}
\end{equation}
The first term captures the current update's direct dependence on parameters; the second propagates dependence through earlier states. Retaining the latter path allows later-view geometric supervision to refine earlier cache writes.

We use StreamVGGT's released losses~\cite{streamvggt} with the weights in \cref*{eq:qad}. These include the 9D camera L1 loss, scale/shift alignment, validity masks, and student-confidence weighting with $-0.1\log(\mathrm{confidence})$ regularization.

\section{Packed-Cache Execution}
\label{app:deployment}
\subsection{Fused Reads and Paired Writes}
\label{app:fused_execution}
\begin{figure}[!htbp]
\centering
\includegraphics[width=\linewidth]{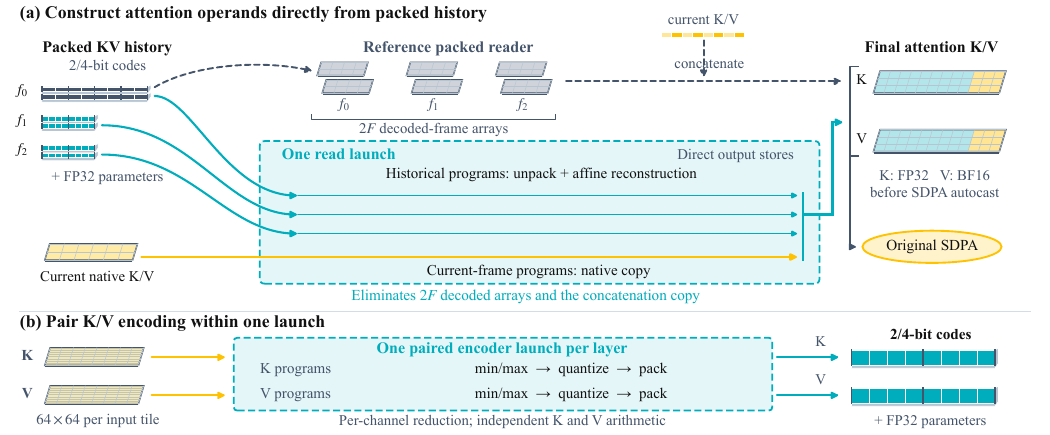}
\caption{\textbf{Fused reads and paired writes.} (a) One launch reconstructs $F$ historical frames and copies current native K/V directly into final attention operands, eliminating $2F$ decoded arrays and concatenation. Dashed outlines mark launch boundaries; SDPA remains outside. (b) Independent K and V programs share one encoder launch per layer, fusing per-channel reduction, quantization, and packing.}
\label{fig:kernel_layout}
\end{figure}

The packed-cache implementation in \cref*{sec:deployment} constructs attention operands without intermediate history materialization (\cref{fig:kernel_layout}). One read launch reconstructs historical K/V from mixed 2/4-bit codes and FP32 affine parameters, while a separate program branch copies current native-precision K/V. Both branches write directly into the final contiguous attention operands. With $F$ historical frames, this removes $2F$ per-frame decoded arrays and the subsequent concatenation. Original SDPA and its BF16 autocast remain outside the fused kernel.

Q and K share a rotation launch after normalization and 2D RoPE; V remains unrotated. Each layer stages current K/V after its FFN. Once the backbone completes, one paired encoder launch per layer performs per-channel range reduction, affine quantization, and bit packing before geometry prediction. Each program processes K or V independently within the shared launch, while previously packed history is reused.

\subsection{Kernel Organization and Numerical Agreement}
\label{app:kernel_correctness}

\begin{table}[!htbp]
\caption{\textbf{Profiled single-request kernel configurations.} $H,N,D,F$ denote heads, tokens per view, head dimension, and historical frames. Profiling uses $H=16$, $N=1{,}004$, $D=64$, and four warps per program, with floating-point contraction disabled.}
\label{tab:kernel_organization}
\centering\small
\renewcommand{\arraystretch}{1.15}
\begin{tabularx}{\linewidth}{@{}p{0.24\linewidth}Xl@{}}
\toprule
Operation & Work per program & Launch grid\\
\midrule
Paired Q/K rotation & 1,024 Q or K values; six FP32 butterfly stages and normalization by $1/8$ & $(\lceil HND/1024\rceil,2)$\\
Fused history read & 1,024 positions in one frame/head; both K and V & $(\lceil ND/1024\rceil,H,F+1)$\\
Paired K/V encoding & 64 tokens $\times$ 64 channels from K or V; per-channel reduction, quantization, and packing & $(H,\lceil N/64\rceil,2)$\\
\bottomrule
\end{tabularx}
\end{table}

For the configuration in \cref{tab:kernel_organization}, Q/K are FP32 and V is BF16 before the original BF16 SDPA autocast. Correctness is checked against unfused reference operations under the same quantization policy. Eight cases cross 65/1,004 tokens, 2/4 bits, and random/constant values, all with noncontiguous inputs; 65 tokens also test an incomplete 64-token group. The encoders produce identical packed codes, FP32 affine parameters, and physical bytes, while paired Q/K rotation matches the reference transform. Three history cases verify identical attention operands, SDPA outputs, and current-write payloads. These checks establish agreement with the quantized reference on the tested paths.

Runtime checks additionally cover prediction shapes, FP32 output dtypes, and request resets. For batched inference, image encoding, transformer blocks, and attention are batched, while camera, depth, and point heads retain their original per-request execution shapes. Each request maintains separate KV and camera state. Complete-inference measurements and kernel cost attribution are reported in \cref{app:resources}.

\section{Complete Geometric Evaluation}
\label{app:results}
\subsection{Evaluation Protocol and Baselines}
\label{app:protocol}
\label{app:configuration_ids}

\begin{table}[!htbp]
\caption{\textbf{Evaluation subsets and primary metrics.} Counts exclude training clips and diagnostic prefixes.}
\label{tab:protocol}
\centering\small
\renewcommand{\arraystretch}{1.08}
\begin{tabularx}{\linewidth}{@{}llrrX@{}}
\toprule
Task & Dataset & Sequences & Frames & Primary metrics\\
\midrule
Depth & Sintel~\cite{sintel} & 23 & 1064 & AbsRel, RMSE, $\delta_1$\\
 & KITTI~\cite{kitti} & 13 & 1269 & AbsRel, RMSE, $\delta_1$\\
 & Bonn~\cite{bonn} & 5 & 550 & AbsRel, RMSE, $\delta_1$\\
Pose & TUM RGB-D~\cite{tum} & 8 & 720 & ATE, translation RPE, rotation RPE\\
Reconstruction & NeuralRGBD~\cite{nrgbd} & 9 & 28 & Accuracy, completeness, normal consistency\\
\bottomrule
\end{tabularx}
\end{table}

\Cref{tab:protocol} summarizes the fixed evaluation subsets. Depth evaluation follows the original scale conventions and validity masks, aggregating over valid pixels. Alongside AbsRel and RMSE, we report SqRel, LogRMSE, and threshold accuracies $\delta_i$, the percentage of valid pixels satisfying $\max(d/\widehat d,\widehat d/d)<1.25^i$, where $d$ and $\widehat d$ denote ground-truth and predicted depth. TUM evaluation uses sequence-level $\mathrm{Sim}(3)$ alignment and adjacent-frame RPE, averaging sequence RMSEs; translation is measured in meters and rotation in degrees. We aggregate point-head reconstruction metrics by arithmetic means across scenes, reporting mean/median distances and both directional normal-consistency components. All methods use the same CPU evaluator and metric definitions; equivalence to every upstream pipeline has not been independently established.

For KIVI and QuaRot, we use the affine RTN grid in \cref*{eq:affine}, with 64-value groups and FP32 metadata. KIVI quantizes keys per channel and values per token. QuaRot uses paired Q/K rotation after normalization and RoPE, offline V/O folding, and per-token K/V groups. Both baselines quantize all historical tokens without precision protection or a native-precision residual window; current-frame K/V, weights, and other activations retain native precision. The no-adaptation control uses pretrained weights under the QuantStream3D precision policy.

Pruning comparisons preserve each method's native selection schedule: InfiniteVGGT selects before attention and XStreamVGGT after attention, while FrameVGGT uses its geometry-based selector with a 65,536-token capacity. Both the pruning-only and QuantStream3D configurations disable XStreamVGGT's original KV quantization.

\subsection{Additional Benchmark Results and QAD Ablations}
\label{app:complete_results}
\label{app:standalone_ablation}
We extend the standalone comparison to Sintel and KITTI for both backbones (\cref{tab:standalone_depth}). On Sintel, QuantStream3D leads the quantized baselines on all three primary metrics for both backbones, while StreamVGGT surpasses Original on all seven reported metrics. On KITTI, it again achieves the best quantized primary metrics for STream3R, including lower RMSE than Original. StreamVGGT also improves all three primary KITTI metrics over Original, although KIVI leads the quantized methods on these metrics.
\begin{table}[!htbp]
\caption{\textbf{Depth estimation on Sintel and KITTI.} Ours denotes QuantStream3D. RMSE is in meters; threshold accuracies are percentages.}
\label{tab:standalone_depth}
\centering\small
\renewcommand{\arraystretch}{1.08}
\setlength{\tabcolsep}{1.5pt}
\begin{tabularx}{\linewidth}{l*{2}{Y{1.15}Y{1.10}Y{0.65}Y{1.10}Y{1.70}Y{0.65}Y{0.65}}}
\toprule
\multicolumn{15}{c}{\textbf{StreamVGGT}}\\
 & \multicolumn{7}{c}{Sintel} & \multicolumn{7}{c}{KITTI}\\
\cmidrule(lr){2-8}\cmidrule(lr){9-15}
Method & AbsRel$\downarrow$ & RMSE$\downarrow$ & $\delta_1$$\uparrow$ & SqRel$\downarrow$ & LogRMSE$\downarrow$ & $\delta_2$$\uparrow$ & $\delta_3$$\uparrow$ & AbsRel$\downarrow$ & RMSE$\downarrow$ & $\delta_1$$\uparrow$ & SqRel$\downarrow$ & LogRMSE$\downarrow$ & $\delta_2$$\uparrow$ & $\delta_3$$\uparrow$\\
\midrule
Original & 0.328 & 3.813 & 65.0 & 1.487 & 0.372 & 82.0 & 89.9 & 0.173 & 4.969 & 72.3 & 0.934 & 0.217 & 94.4 & 99.3\\
KIVI & 0.326 & 3.826 & 63.2 & 1.436 & 0.374 & 81.9 & 90.0 & 0.161 & 4.869 & 76.2 & 0.851 & 0.206 & 95.6 & 99.2\\
QuaRot & 0.333 & 3.850 & 63.6 & 1.482 & 0.377 & 81.7 & 89.7 & 0.170 & 4.977 & 73.2 & 0.910 & 0.215 & 94.8 & 99.3\\
\rowcolor{black!6}
Ours & 0.318 & 3.726 & 65.8 & 1.413 & 0.359 & 83.2 & 90.3 & 0.171 & 4.877 & 72.9 & 0.905 & 0.214 & 94.5 & 99.3\\
\bottomrule
\end{tabularx}
\par\smallskip
\begin{tabularx}{\linewidth}{l*{2}{Y{1.15}Y{1.10}Y{0.65}Y{1.10}Y{1.70}Y{0.65}Y{0.65}}}
\toprule
\multicolumn{15}{c}{\textbf{STream3R}}\\
 & \multicolumn{7}{c}{Sintel} & \multicolumn{7}{c}{KITTI}\\
\cmidrule(lr){2-8}\cmidrule(lr){9-15}
Method & AbsRel$\downarrow$ & RMSE$\downarrow$ & $\delta_1$$\uparrow$ & SqRel$\downarrow$ & LogRMSE$\downarrow$ & $\delta_2$$\uparrow$ & $\delta_3$$\uparrow$ & AbsRel$\downarrow$ & RMSE$\downarrow$ & $\delta_1$$\uparrow$ & SqRel$\downarrow$ & LogRMSE$\downarrow$ & $\delta_2$$\uparrow$ & $\delta_3$$\uparrow$\\
\midrule
Original & 0.265 & 3.083 & 70.6 & 1.226 & 0.322 & 85.7 & 93.1 & 0.079 & 3.294 & 94.9 & 0.344 & 0.116 & 98.9 & 99.7\\
KIVI & 0.278 & 3.111 & 67.9 & 1.170 & 0.333 & 84.9 & 92.5 & 0.113 & 4.176 & 87.7 & 0.591 & 0.151 & 98.3 & 99.6\\
QuaRot & 0.277 & 3.177 & 67.7 & 1.300 & 0.333 & 85.7 & 93.0 & 0.102 & 3.869 & 91.5 & 0.491 & 0.139 & 98.5 & 99.6\\
\rowcolor{black!6}
Ours & 0.272 & 3.095 & 68.8 & 1.066 & 0.317 & 86.1 & 91.8 & 0.080 & 3.229 & 95.6 & 0.327 & 0.117 & 99.0 & 99.6\\
\bottomrule
\end{tabularx}
\end{table}

To isolate Geometry-Aligned QAD, we compare pretrained and adapted weights under the same precision policy and KV storage (\cref{tab:ablation_depth}). QAD improves all three primary depth metrics across the three datasets for StreamVGGT, with additional gains on KITTI and Bonn for STream3R. Relative to the no-QAD control, Bonn AbsRel decreases by $7.2\%$ for StreamVGGT and $10.3\%$ for STream3R, with both adapted models surpassing Original. These controlled comparisons demonstrate additional geometric accuracy gains at unchanged cache storage, while the mixed STream3R Sintel results show that the gains vary across backbones and datasets.
\begin{table}[!htbp]
\caption{\textbf{Geometry-Aligned QAD on depth.} Shading denotes QuantStream3D; both quantized variants share the precision policy and KV storage. RMSE is in meters and threshold accuracies are percentages.}
\label{tab:ablation_depth}
\centering\small
\renewcommand{\arraystretch}{1.08}
\setlength{\tabcolsep}{2pt}
\begin{tabularx}{\linewidth}{@{}lll*{7}{>{\raggedleft\arraybackslash}X}}
\toprule
Dataset & Backbone & Variant & AbsRel$\downarrow$ & RMSE$\downarrow$ & $\delta_1$$\uparrow$ & SqRel$\downarrow$ & LogRMSE$\downarrow$ & $\delta_2$$\uparrow$ & $\delta_3$$\uparrow$\\
\midrule
\multirow{6}{*}{Sintel} & \multirow{3}{*}{StreamVGGT} & Original & 0.3280 & 3.8127 & 65.0 & 1.4868 & 0.3720 & 82.0 & 89.9\\
 &  & \noqad & 0.3289 & 3.8145 & 65.0 & 1.5043 & 0.3722 & 81.8 & 89.9\\
 &  & \cellcolor{black!6}\method & \cellcolor{black!6}0.3176 & \cellcolor{black!6}3.7260 & \cellcolor{black!6}65.8 & \cellcolor{black!6}1.4129 & \cellcolor{black!6}0.3587 & \cellcolor{black!6}83.2 & \cellcolor{black!6}90.3\\
\cmidrule(l){2-10}
 & \multirow{3}{*}{STream3R} & Original & 0.2653 & 3.0825 & 70.6 & 1.2259 & 0.3216 & 85.7 & 93.1\\
 &  & \noqad & 0.2672 & 3.0954 & 70.3 & 1.2332 & 0.3227 & 85.3 & 93.1\\
 &  & \cellcolor{black!6}\method & \cellcolor{black!6}0.2722 & \cellcolor{black!6}3.0952 & \cellcolor{black!6}68.8 & \cellcolor{black!6}1.0660 & \cellcolor{black!6}0.3173 & \cellcolor{black!6}86.1 & \cellcolor{black!6}91.8\\
\midrule
\multirow{6}{*}{KITTI} & \multirow{3}{*}{StreamVGGT} & Original & 0.1726 & 4.9688 & 72.3 & 0.9344 & 0.2171 & 94.4 & 99.3\\
 &  & \noqad & 0.1746 & 5.0079 & 71.7 & 0.9461 & 0.2191 & 94.4 & 99.2\\
 &  & \cellcolor{black!6}\method & \cellcolor{black!6}0.1712 & \cellcolor{black!6}4.8770 & \cellcolor{black!6}72.9 & \cellcolor{black!6}0.9052 & \cellcolor{black!6}0.2137 & \cellcolor{black!6}94.5 & \cellcolor{black!6}99.3\\
\cmidrule(l){2-10}
 & \multirow{3}{*}{STream3R} & Original & 0.0790 & 3.2939 & 94.9 & 0.3442 & 0.1165 & 98.9 & 99.7\\
 &  & \noqad & 0.0800 & 3.3045 & 95.0 & 0.3484 & 0.1174 & 98.9 & 99.7\\
 &  & \cellcolor{black!6}\method & \cellcolor{black!6}0.0800 & \cellcolor{black!6}3.2287 & \cellcolor{black!6}95.6 & \cellcolor{black!6}0.3268 & \cellcolor{black!6}0.1168 & \cellcolor{black!6}99.0 & \cellcolor{black!6}99.6\\
\midrule
\multirow{6}{*}{Bonn} & \multirow{3}{*}{StreamVGGT} & Original & 0.0585 & 0.2601 & 97.2 & 0.0344 & 0.1142 & 98.4 & 99.0\\
 &  & \noqad & 0.0589 & 0.2597 & 97.2 & 0.0345 & 0.1141 & 98.4 & 99.0\\
 &  & \cellcolor{black!6}\method & \cellcolor{black!6}0.0547 & \cellcolor{black!6}0.2456 & \cellcolor{black!6}97.5 & \cellcolor{black!6}0.0335 & \cellcolor{black!6}0.1075 & \cellcolor{black!6}98.5 & \cellcolor{black!6}99.1\\
\cmidrule(l){2-10}
 & \multirow{3}{*}{STream3R} & Original & 0.0697 & 0.2700 & 95.2 & 0.0451 & 0.1325 & 98.4 & 98.9\\
 &  & \noqad & 0.0689 & 0.2676 & 96.1 & 0.0444 & 0.1309 & 98.4 & 98.9\\
 &  & \cellcolor{black!6}\method & \cellcolor{black!6}0.0618 & \cellcolor{black!6}0.2568 & \cellcolor{black!6}97.4 & \cellcolor{black!6}0.0432 & \cellcolor{black!6}0.1192 & \cellcolor{black!6}98.4 & \cellcolor{black!6}98.9\\
\bottomrule
\end{tabularx}
\end{table}

\subsection{Additional Post-Pruning Reconstruction Results}
\label{app:composition}
\Cref{tab:reconstruction_all} extends the NeuralRGBD comparison in \cref*{sec:composition} with median distances and additional normal-consistency measures. Adding QuantStream3D improves mean accuracy, completeness, and normal consistency for all three pruning methods relative to pruning alone. Auxiliary metrics show mixed changes across pruning methods.
\begin{table}[!htbp]
\caption{\textbf{Reconstruction after native pruning on NeuralRGBD.} Original is unpruned and unquantized; Ours denotes QuantStream3D. Distances are in meters; normal consistency is unitless.}
\label{tab:reconstruction_all}
\centering\small
\renewcommand{\arraystretch}{1.08}
\setlength{\tabcolsep}{2pt}
\begin{tabularx}{\linewidth}{@{}ll*{8}{>{\raggedleft\arraybackslash}X}}
\toprule
Pruning method & Configuration & Acc.$\downarrow$ & Comp.$\downarrow$ & NC$\uparrow$ & Acc. med.$\downarrow$ & Comp. med.$\downarrow$ & NC med.$\uparrow$ & NC1$\uparrow$ & NC2$\uparrow$\\
\midrule
Unpruned StreamVGGT & Original & 0.084 & 0.079 & 0.861 & 0.044 & 0.039 & 0.986 & 0.862 & 0.860\\
\cmidrule(l){1-10}
\multirow{2}{*}{InfiniteVGGT} & Prune only & 0.084 & 0.079 & 0.861 & 0.044 & 0.039 & 0.986 & 0.862 & 0.860\\
 & \cellcolor{black!6}Prune + Ours & \cellcolor{black!6}0.074 & \cellcolor{black!6}0.066 & \cellcolor{black!6}0.867 & \cellcolor{black!6}0.043 & \cellcolor{black!6}0.033 & \cellcolor{black!6}0.990 & \cellcolor{black!6}0.876 & \cellcolor{black!6}0.858\\
\cmidrule(l){1-10}
\multirow{2}{*}{XStreamVGGT} & Prune only & 0.084 & 0.076 & 0.866 & 0.052 & 0.035 & 0.990 & 0.865 & 0.867\\
 & \cellcolor{black!6}Prune + Ours & \cellcolor{black!6}0.074 & \cellcolor{black!6}0.063 & \cellcolor{black!6}0.869 & \cellcolor{black!6}0.055 & \cellcolor{black!6}0.034 & \cellcolor{black!6}0.991 & \cellcolor{black!6}0.875 & \cellcolor{black!6}0.863\\
\cmidrule(l){1-10}
\multirow{2}{*}{FrameVGGT} & Prune only & 0.084 & 0.079 & 0.861 & 0.044 & 0.039 & 0.986 & 0.862 & 0.860\\
 & \cellcolor{black!6}Prune + Ours & \cellcolor{black!6}0.074 & \cellcolor{black!6}0.066 & \cellcolor{black!6}0.867 & \cellcolor{black!6}0.043 & \cellcolor{black!6}0.033 & \cellcolor{black!6}0.990 & \cellcolor{black!6}0.876 & \cellcolor{black!6}0.858\\
\bottomrule
\end{tabularx}
\end{table}

\section{Efficiency and Mechanism Ablations}
\label{app:resources}
\subsection{Complete-Inference Measurements}
\label{app:efficiency}
We profile complete StreamVGGT inference on one NVIDIA H20 with fixed pretrained weights, original SDPA, BF16 autocast, and the original FP32 output format. One full-workload warm-up precedes three measured rounds. Timing runs from inference-API entry until all outputs are CPU-ready, excluding preprocessing, host-to-device transfer, and disk serialization. These profiles isolate backend execution; geometric accuracy is evaluated with the adapted checkpoints. At batch size one, each round processes eight sequential 90-view TUM requests (720 views) under a 60-GiB allocator budget (\cref{tab:single_efficiency}). Allocated and reserved memory peaks cover the measured rounds; persistent KV storage counts each allocation once, including metadata and padding. Peak allocation also includes weights, decoded operands, activations, workspaces, and camera state, so KV compression need not produce proportional peak-memory savings.

\par\smallskip\noindent\begin{minipage}{\linewidth}
\centering\small\captionsetup{hypcap=false}
\setlength{\tabcolsep}{7pt}
\captionof{table}{\textbf{Batch-size-one inference under a 60-GiB allocator budget.} Each round contains eight sequential 90-view requests; times are medians of three rounds.}
\label{tab:single_efficiency}
\begin{tabularx}{\textwidth}{lY{1.25}Y{0.65}Y{1.05}Y{1.25}Y{0.80}}
\toprule
KV policy & 720-view time (s)$\downarrow$ & Frames/s$\uparrow$ & Peak alloc. (GiB)$\downarrow$ & Peak reserved (GiB)$\downarrow$ & KV (GiB)$\downarrow$\\
\midrule
Original & 175.73 & 4.10 & 15.40 & 41.84 & 8.27\\
\rowcolor{black!6}
\packedbackend & 135.14 & 5.33 & 8.26 & 26.74 & 1.64\\
\bottomrule
\end{tabularx}
\end{minipage}\par\smallskip
Batch scaling uses the same H20 under a separate 120-GiB allocator budget, cycling the eight TUM sequences as needed with independent histories and request resets (\cref{tab:efficiency_all}). Throughput is total completed frames divided by median round time; request latency is batch-completion time and is not divided by batch size. The largest completed batches are 5 for Original and 27 for packed KV, bounded by adjacent OOM tests at 6 and 28. Peak throughput is $4.700$ frames/s at Original batch 4 and $7.835$ frames/s at packed batch 26, a $1.67\times$ improvement.

\begingroup\small
\renewcommand{\arraystretch}{1.08}
\setlength{\tabcolsep}{5pt}
\setlength{\LTpre}{4pt}\setlength{\LTpost}{6pt}\setlength{\LTcapwidth}{\textwidth}
\begin{longtable}{@{}rrrrrrr@{}}
\caption{\textbf{Batch scaling under the 120-GiB budget.} Throughput counts all completed frames; request latency gives the median [min, max] batch-completion time over three rounds. Peaks exclude correctness checks; KV reports unique persistent allocations.}\label{tab:efficiency_all}\\
\toprule
Batch & \shortstack{Frames\\per round} & \shortstack{Throughput\\(frames/s)} & \shortstack{90-view request latency\\median [min, max] (s)} & \shortstack{Peak alloc.\\(GiB)} & \shortstack{Peak reserved\\(GiB)} & \shortstack{Unique KV\\(GiB)}\\
\midrule
\endfirsthead
\multicolumn{7}{l}{\tablename~\thetable\ (continued)}\\
\toprule
Batch & \shortstack{Frames\\per round} & \shortstack{Throughput\\(frames/s)} & \shortstack{90-view request latency\\median [min, max] (s)} & \shortstack{Peak alloc.\\(GiB)} & \shortstack{Peak reserved\\(GiB)} & \shortstack{Unique KV\\(GiB)}\\
\midrule
\endhead
\midrule\multicolumn{7}{r}{Continued on the next page}\\\endfoot
\bottomrule\endlastfoot
\multicolumn{7}{l}{\textbf{Original (BF16 KV)}}\\*
1 & 720 & 4.10 & 21.94 [21.91, 21.99] & 15.40 & 39.76 & 8.27\\*
2 & 720 & 4.36 & 41.28 [41.24, 41.30] & 26.17 & 87.13 & 16.55\\*
4 & 720 & 4.70 & 76.59 [76.48, 76.69] & 47.45 & 119.67 & 33.09\\*
5 & 900 & 4.69 & 95.98 [95.97, 95.99] & 58.11 & 117.16 & 41.36\\
\midrule
\multicolumn{7}{l}{\packedbackend}\\*
1 & 720 & 5.32 & 16.93 [16.86, 17.00] & 8.26 & 26.72 & 1.64\\*
2 & 720 & 6.17 & 29.17 [29.07, 29.27] & 11.89 & 52.04 & 3.27\\*
4 & 720 & 7.11 & 50.68 [50.54, 50.82] & 18.95 & 90.71 & 6.55\\*
8 & 720 & 7.43 & 96.85 [96.82, 96.91] & 33.17 & 67.93 & 13.10\\*
16 & 1440 & 7.68 & 187.42 [187.41, 187.50] & 61.35 & 94.90 & 26.20\\*
24 & 2160 & 7.77 & 277.86 [277.82, 278.06] & 89.96 & 109.79 & 39.29\\*
26 & 2340 & 7.83 & 298.67 [298.63, 298.73] & 97.05 & 107.28 & 42.57\\*
27 & 2430 & 7.81 & 310.95 [310.50, 311.94] & 100.60 & 113.10 & 44.20\\
\end{longtable}
\endgroup

\subsection{Kernel Cost Attribution}
\label{app:kernel_ablation}
We profile synthetic StreamVGGT tensors on one H20 using PyTorch 2.8.0, CUDA 12.8, and Triton 3.4.0. Inputs have one request, 16 heads, head dimension 64, and 1,004 current-view tokens. Q/K are FP32 and V is BF16, with original SDPA under BF16 autocast. Global attention in layer 0 uses 36 or 89 historical frames, with the first protected at 4 bits and others at 2; layer 14 uses 36 all-4-bit frames. The timed pipeline covers 64-dimensional Hadamard Q/K rotation, history reconstruction with current-K/V appending, original SDPA, and current K/V encoding at fixed history lengths. Each control disables history-read fusion, Q/K pairing, or K/V pairing individually while holding inputs and precision fixed. Separate K/V encoding retains fused range reduction, rounding, and packing within each operand, isolating the effect of pairing. After JIT compilation, Q/K rotation and K/V encoding use 12 warm-ups and 30 calls per round; history reads and the pipeline use five warm-ups and 15 calls. Five interleaved rounds randomize variant order, using synchronized CUDA-event timing that includes format validation and allocation. Numerical agreement is documented in \cref{app:kernel_correctness}.

Our fused reader drives the pipeline gain by reconstructing packed history and appending current K/V directly into final attention operands, eliminating per-frame decoded buffers and a separate concatenation step. This reduces isolated history-read latency by up to a factor of $23.63$ and approximately halves temporary allocation (\cref{tab:kernel_components}). Within the profiled pipeline, history-read fusion provides $2.54\times$ and $2.63\times$ speedups for 36 and 89 historical frames, respectively. At protected layer 14, pipeline latency decreases from $7.112$ to $2.800$ ms, showing that the benefit also extends to 4-bit history. We also pair Q and K rotation within one kernel launch and encode K and V together, reducing the number of kernel launches. Q/K pairing achieves a $1.40\times$ isolated-stage speedup and K/V pairing up to $1.13\times$, while paired and separate pipeline timing ranges overlap. Complete-inference gains are measured separately in \cref{app:efficiency}.
\begin{table}[!htbp]
\caption{\textbf{Isolated kernel costs and incremental allocation.} GPU latency is the median [minimum, maximum] over five interleaved rounds. Temporary allocation is measured above persistent inputs and includes output buffers. History-read fusion removes decoded-frame intermediates; pairing preserves output allocation.}
\label{tab:kernel_components}
\centering\small
\renewcommand{\arraystretch}{1.10}
\setlength{\tabcolsep}{4pt}
\begin{tabular*}{\textwidth}{@{\extracolsep{\fill}}lrrrrr@{}}
\toprule
Component & \shortstack{Separate / reference\\GPU time (ms)} & \shortstack{Paired / fused\\GPU time (ms)} & Gain & \shortstack{Reference temp.\\(MiB)} & \shortstack{Fused temp.\\(MiB)}\\
\midrule
Paired Q/K rotation & 0.041 [0.041, 0.042] & 0.030 [0.029, 0.030] & $1.40\times$ & 7.84 & 7.84\\
Paired K/V, 2-bit & 0.111 [0.108, 0.112] & 0.098 [0.097, 0.100] & $1.13\times$ & 0.75 & 0.75\\
Paired K/V, 4-bit & 0.110 [0.109, 0.122] & 0.098 [0.097, 0.099] & $1.12\times$ & 1.25 & 1.25\\
\midrule
History read, $F=36$ & 4.205 [4.178, 4.306] & 0.213 [0.201, 0.227] & $19.76\times$ & 434.79 & 218.56\\
History read, $F=89$ & 10.360 [10.347, 10.588] & 0.438 [0.425, 0.454] & $23.63\times$ & 1063.73 & 529.46\\
History read, $F=36$, Layer 14 & 4.198 [4.179, 4.267] & 0.215 [0.201, 0.239] & $19.53\times$ & 434.79 & 218.56\\
\bottomrule
\end{tabular*}
\end{table}

\subsection{Targeted Mechanism Ablations}
\label{app:ablation_protocol}
The $2\times2$ precision-placement study crosses protected frames 0/1 with global-attention layers 13/14 under the same 2-bit background. All four policies share pretrained weights, inputs, token order, the Q/K basis, unrotated V, and per-channel groups of 64. For $L=24$ layers and $T\ge2$ equal-shape views, each union protects $L+T-1$ writes. Let $S_b$ count all codes, affine parameters, and padding in one $b$-bit K/V write. Total physical storage is
\begin{equation}
 M=(L+T-1)S_4+(LT-L-T+1)S_2.
\label{eq:placement_bytes}
\end{equation}
All four policies therefore have identical final KV bytes within each sequence, although intermediate cache sizes and peak memory may differ. Protecting layer 14 improves all three TUM trajectory metrics for either protected frame (\cref{tab:placement_full}). On NeuralRGBD, the preferred protected frame for completeness reverses between layers 13 and 14. Precision locations thus interact: allocating the same number of bits at different positions changes geometric accuracy.

\begin{table}[!htbp]
\caption{\textbf{Precision placement at equal final KV storage.} Frozen StreamVGGT with a 2-bit background; frame 0/layer 14 is our policy. Each row protects the union of one frame and one layer. Distances are in meters and rotational RPE in degrees.}
\label{tab:placement_full}
\centering\small
\setlength{\tabcolsep}{6pt}
\renewcommand{\arraystretch}{1.05}
\begin{tabular}{lrrrrrr}
\toprule
 & \multicolumn{3}{c}{TUM RGB-D} & \multicolumn{3}{c}{NeuralRGBD}\\
\cmidrule(lr){2-4}\cmidrule(lr){5-7}
Protection & ATE$\downarrow$ & RPE-t$\downarrow$ & RPE-r$\downarrow$ & Acc.$\downarrow$ & Comp.$\downarrow$ & NC$\uparrow$\\
\midrule
Frame 0 / Layer 14 & 0.0272 & 0.0127 & 0.3215 & 0.0827 & 0.0764 & 0.8604\\
Frame 1 / Layer 14 & 0.0275 & 0.0131 & 0.3213 & 0.0804 & 0.0753 & 0.8595\\
Frame 0 / Layer 13 & 0.0347 & 0.0175 & 0.3567 & 0.0835 & 0.0779 & 0.8605\\
Frame 1 / Layer 13 & 0.0352 & 0.0188 & 0.3708 & 0.0865 & 0.0789 & 0.8521\\
\bottomrule
\end{tabular}
\end{table}

Adapted variants share initialization, teacher, losses, data order, optimizer schedule, cache policy, and 2,000 updates. Geometry-head adaptation updates the camera, depth, and point heads; transformer-block adaptation updates complete frame- and global-attention blocks, including FFNs and normalization. Joint adaptation updates both; the image encoder and seed tokens remain frozen. All variants generate their own quantized histories under the same graph boundaries.

Joint adaptation leads all three NeuralRGBD metrics (\cref{tab:scope_complete}), supporting adaptation of both history processing and geometric prediction. Transformer-block adaptation alone gives the lowest TUM translation errors, and geometry-head adaptation alone the lowest rotational RPE. These results complement the local correction-capacity analysis in \cref*{app:joint_scope}.

\par\smallskip\noindent\begin{minipage}{\linewidth}
\captionsetup{hypcap=false}
\captionof{table}{\textbf{Adaptation scope with a fixed cache policy.} Adapted variants share 2,000 updates; the no-adaptation control retains pretrained weights. Distances are in meters and rotational RPE in degrees. Shading denotes joint adaptation.}
\label{tab:scope_complete}
\centering\small
\setlength{\tabcolsep}{7pt}
\renewcommand{\arraystretch}{1.08}
\begin{tabular}{lrrrrrr}
\toprule
 & \multicolumn{3}{c}{TUM RGB-D} & \multicolumn{3}{c}{NeuralRGBD}\\
\cmidrule(lr){2-4}\cmidrule(lr){5-7}
Update scope & ATE$\downarrow$ & RPE-t$\downarrow$ & RPE-r$\downarrow$ & Acc.$\downarrow$ & Comp.$\downarrow$ & NC$\uparrow$\\
\midrule
No adaptation & 0.0272 & 0.0127 & 0.3215 & 0.0827 & 0.0764 & 0.8604\\
Geometry heads only & 0.0274 & 0.0127 & 0.3209 & 0.0823 & 0.0760 & 0.8597\\
Transformer blocks only & 0.0232 & 0.0119 & 0.3238 & 0.0818 & 0.0753 & 0.8660\\
\rowcolor{black!6}
Joint (\method) & 0.0264 & 0.0126 & 0.3220 & 0.0740 & 0.0661 & 0.8668\\
\bottomrule
\end{tabular}
\end{minipage}\par\smallskip

\end{document}